\documentclass[10pt,twocolumn,letterpaper]{article}

\usepackage{iccv}
\usepackage{times}
\usepackage{epsfig}
\usepackage{graphicx}
\usepackage{amsmath}
\usepackage{amssymb}
\usepackage{esvect}

\usepackage{url}
\usepackage{microtype}
\usepackage{graphicx}
\usepackage{subfigure}
\usepackage{booktabs} 
\usepackage{xcolor}
\usepackage{amsthm}
\usepackage{multirow}
\usepackage{diagbox}
\usepackage{arydshln}
\usepackage{mathtools}
\usepackage{soul}
\soulregister\cite7
\soulregister\gls7
\soulregister\etal7

\usepackage{booktabs}
\usepackage{float}
\usepackage{hhline}
\usepackage{tabularx}
\newcolumntype{Y}{>{\centering\arraybackslash}X}
\newcolumntype{s}{>{\hsize=.8\hsize}Y}
\newcolumntype{t}{>{\hsize=.6\hsize}Y}

\newcolumntype{?}{!{\vrule width 1pt}}
\usepackage{multirow}
\usepackage{makecell}
\usepackage{stfloats}
\usepackage{wrapfig}
\usepackage{dsfont}

\newcommand*\samethanks[1][\value{footnote}]{\footnotemark[#1]}

\usepackage[breaklinks=true,bookmarks=false]{hyperref}

\iccvfinalcopy 

\def\iccvPaperID{7605} 

\ificcvfinal\fi

\begin{document}



\title{A Hierarchical Transformation-Discriminating Generative\\ Model for Few Shot Anomaly Detection}

\author {Shelly Sheynin$^{1}$\thanks{Equal contribution} \qquad Sagie Benaim$^{1}$\samethanks \qquad Lior Wolf$^{1,2}$ \\
$^1$\normalsize The School of Computer Science, Tel Aviv University\\
$^2$\normalsize Facebook AI Research
}
\maketitle
\ificcvfinal\thispagestyle{empty}\fi
\begin{abstract}
Anomaly detection, the task of identifying unusual samples in data, 
often relies on a large set of training samples. 
In this work, we consider the setting of few-shot anomaly detection in images, where only a few images are given at training. We devise a hierarchical generative model that captures the multi-scale patch distribution of each training image. We further enhance the representation of our model by using image transformations and optimize scale-specific patch-discriminators to  distinguish between real and fake patches of the image, as well as between different transformations applied to those patches. The anomaly score is obtained by aggregating the patch-based votes of the correct transformation across scales and image regions. 
We demonstrate the superiority of our method on both the one-shot and few-shot settings, on the datasets of Paris, CIFAR10, MNIST and FashionMNIST as well as in the setting of defect detection on MVTec. In all cases, our method outperforms the recent baseline methods. 
\end{abstract}

\section{Introduction}
\label{sec:intro}




\textit{Anomaly detection}~\cite{chandola2009anomaly, aggarwal2015outlier} is the task of detecting unusual samples in the data. In the typical setting of  \textit{one class classification}~\cite{moya1993one}, given a large collection of samples from the \textit{normal} (non-anomalous) data, the learner is asked to classify novel samples as either normal or anomalous. In this paper, we aim to solve this problem given very few training samples, including the case of a single training sample. Our study is motivated by the scarcity of training samples in many visual domains,  as well as by the human ability to solve this task after observing a very limited number of samples~\cite{fei2006knowledge, fei2006one}. 

Our model relies on two main components. The first component is a hierarchical generative model that captures the internal patch statistics of one or few images at multiple scales. This component follows the recent successes of deep generative models to generate multiple, varied, natural looking images, given a single training image~\cite{singan, gur2020hierarchical, ingan}. We generalize this setting to few images, by adding conditioning on the image index, thus enabling information sharing between multiple training images.
 
Modeling patches at different scales, allows  the detection of anomalies both in global properties, such as color or large-scale structural changes, and in local regions. The multi-scale patch based approach is also useful when considering the task of defect detection, where the anomaly may only manifest in few well localized regions. 

The second component is a \textit{self-supervised learning} task. Recent approaches~\cite{golan2018deep,bergman2020classification,goyal2020drocc} have shown that, in the many-shot case, a classifier trained on a proxy task, such as identifying the transformation applied to the input, accurately captures novel samples that are similar to the training data, and thus can distinguish between normal and anomalous samples. We utilize such a proxy task in the context of our multi-scale generative model. For each scale, the generated patches are transformed according to a predefined set of transformations. A discriminator is asked to distinguish between real and fake samples, as well as between the transformations performed. 

Our method builds these components into a single model. At test time, a sample is considered anomalous, if many of its patches, at all scales, are determined anomalous, according to the learned patch-specific discriminators. In a comprehensive set of the experiments, we demonstrate that the proposed method outperforms recent baselines on anomaly detection benchmarks. This is true for the one-shot, five-shot and ten-shot settings. When the number of training samples increases, our method is shown to improve in performance. In the field of defect detection, we show that our method outperforms, in the few shot setting, the state of the art in localizing the anomalous patch.

\section{Related Work}

\noindent\textbf{Image based anomaly detection}\quad
Image based anomaly detection methods can be divided into the categories of \textit{reconstruction}, \textit{classification} and \textit{distribution} based methods~\cite{ruff2021unifying}. 

\textit{Reconstruction} based methods represent data in a manner whereby normal data can be reconstructed with small error, while anomalous data incurs a high reconstruction error. 


\textit{Classification} based methods attempt to discriminate between data regions of normal data and those of anomalous data. 
Ruff et al.~\cite{ruff2018deep} proposed DeepSVDD, a deep learning variant of SVDD, which maps images to a more meaningful deep feature space. PatchSVDD~\cite{yi2020patch} improved this method, by extending it to a patch-based method using self-supervised learning. Recent \textit{self-supervised} approaches attempt to find a  ``proxy" classification objective, such that classifying normal data based on this objective, allows for a good separation of normal and anomalous data. For instance, Goyal et al.~\cite{goyal2020drocc} 
trains a classifier to distinguish training samples from their perturbations generated adversarially. Golan and El-Yaniv~\cite{golan2018deep} use a set of predefined transformations. They train a classifier to distinguish between the type of transformation performed on normal data and show that such a classifier can be used to distinguish between normal and anomalous data. While our method uses such a proxy objective, it is used within an hierarchical patch based generative setting. This allows the modeling of the patch based distribution of images together with a discriminative capability to model the regions of this distribution. 


\textit{Distribution} based methods model the distribution of normal data. Deep generative models have shown great promise in modeling complicated distributions. In particular, autoencoders and variational autoencoders~\cite{zhai2016deep, an2015variational}, as well as GAN based approaches~\cite{schlegl2017unsupervised} are typically used. 
Nalisnick et al.,~\cite{nalisnick2018deep} however, argue that a generative model alone may not be sufficient to detect out-of-distribution inputs. 

Unlike these approaches, our model benefits from the use of both \textit{distribution} and \textit{classification} based components:
(1) A hierarchical generative model used to model the internal multi-scale patch distribution of a single or few images
and (2) A scale-dependent discriminator, which learns to distinguish between real and fake image patches, as well as between transformations of such patches. 
Our model incorporates these components into a single model.  Unlike previous approaches, this allows for the detection of anomalous samples
in the case where, during training, only a single or a few images are given from the normal class.

\noindent\textbf{Few shot learning}\quad
The use of limited supervision for image classification has been studied extensively \cite{wang2020generalizing, chen2019closer, tian2020rethinking}. Our work relates more closely to anomaly detection works that use limited supervision. Some works \cite{pang2018learning, pang2019deep} consider the setting where a limited number of samples is given from the anomalous classes, but many samples are given from the normal class. 
Kozerawski et al.,~\cite{kozerawski2018clear} use large labelled dataset generated from ImageNet and apply transfer learning. 
Other works \cite{pang2019deep2, pang2020deep} tackle a slightly different problem, where few samples are given from novel anomalous classes.  
Frikha et al.,~\cite{frikha2020few} and Kruspe~\cite{kruspe2019one} consider a meta-learning approach where few samples are given from many classes at training. 
All of these works use additional supervision, not used in our method. 
Our work assumes uses only a limited number of samples from the normal class. 


\noindent\textbf{Learning the Internal Statistics}\quad
Our work is related to recent works that model the internal distribution of images~\cite{Zhou2018,ingan,singan, gur2020hierarchical}. These works present a GAN capable of generating from the internal patch distribution of a single image. However, these methods are used in the context of generation, and are designed on a single image only. 




\begin{figure}
\centering
\begin{tabular}{c}
\includegraphics[width=0.995\linewidth]{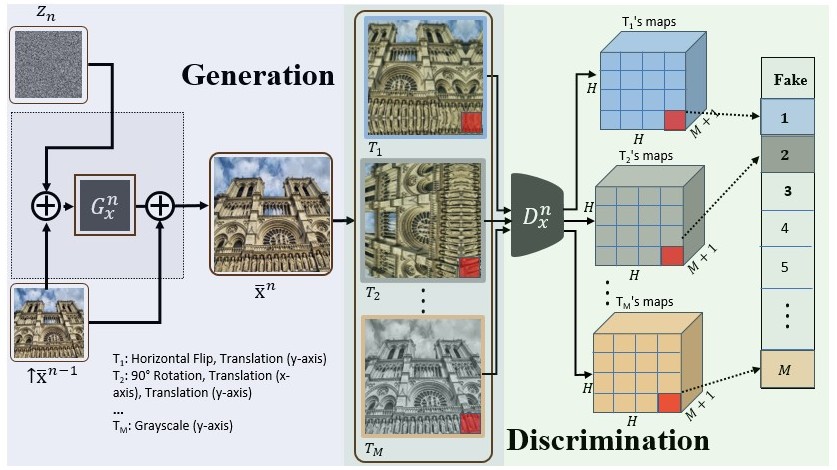} \\
\end{tabular}
\caption{\textbf{Training pipeline at scale $n$ for the one shot setting.} Generator $G_x^n$ receives as input, the upsampled image from previous scale $\overline{x}^{n-1}$ and the noise map $z^n$, and generates a new sample $\overline{x}^{n}$. $\overline{x}^n$ is transformed into the set $\{T_1(\overline{x}^n), \dots, T_M(\overline{x}^n)\}$ of M transformed images and fed into the multi-class discriminator $D^n_x$. For each $T_i$, the result is $M+1$ maps of size $H \times H$. Each $M+1$ sized vector of this $H\times H$ map represents a softmax probability vector over $M+1$ classes for patch $p$ (for instance the red patch at the bottom right). For each such vector, $G_n^x$ attempts to maximize the probability of class $i$ for an image transformed using $T_i$. 
} 
\label{fig:illustration}
\end{figure}

\section{Method}
\label{sec:one_shot}

Let $\mathbb{X}$ be the set of all natural images and let $X \subset \mathbb{X}$ be the subset of images in the class that is defined by the training images, i.e., the normal class. Images residing outside $X$ are considered anomalous. 
We are given an i.i.d randomly selected subset of images $X_k = \{x_1, \dots, x_k \} \subset X$. 
The task is to learn a classifier $C_{X_k}$ from the training set $X_k$, such that $C_{X_k}(x) = 1$ if $x \in X$ and $C_{X_k}(x) = 0$ otherwise. We are interested in the case where $k$ is small. 

\subsection{One Shot Anomaly Detection}

We begin with the case of $k=1$, where the training set contains a single sample $x_1$, denoted as $x$ for brevity. While $x$ alone cannot capture the intra-class variability of the normal class, we argue that the image itself contains informative structure that can help identify the class. An illustration of our training pipeline is provided in Fig.~\ref{fig:illustration}. 

\noindent\textbf{Internal distribution of patches}\quad
Our sample space can be significantly enriched by considering different image patches, at different scales. To this end, we model the distribution of the patches of $x$ at different scales, using a similar pipeline to SinGAN~\cite{singan}. Unlike SinGAN, we use a multi-class discriminator to distinguish between different classes of transformations applied on real and generated samples.

For a given scale $n=0,1,2,\dots,N$, we denote by $p^n_{x}$ the distribution of images that have the same patch distribution at this scale as $x$. In other words, $x$ is composed of many patches at a given distribution and we model the  distribution $p^n_x$ of images that have the same scale-dependent patch distribution. For each scale $n$, the downsampling factor is $r^{N-n}$ for some $r > 1$. Scale $0$ is of the lowest resolution and scale $N$ is of the highest (original) image resolution. A bicubic downsampling is used. 

 At each scale $n$, our method employs a patch GAN~\cite{CycleGAN2017, disc2} for generating samples in $p^n_x$. Each patch-GAN consists of a fully convolutional generator $G^n_x$ and discriminator $D^n_x$ with a receptive field of size $11\times 11$.   
The architecture of $G^n_x$ and $D^n_x$ and training details are outlined below in Sec.~\ref{sec:architecture}. 
Let $x^n$ be $x$, down-sampled by a factor of $r^{N-n}$, and $z^n$ be Gaussian noise of the same dimension and shape as $x^n$. At the coarsest scale, $n=0$, the output of $G^{n}_x$ is defined as (the overline is used to denote generated images):
\begin{align}
\overline{x}^{0} = G^{0}_x(z^{0}) \label{eq:uncon1}
\end{align}
Moving to finer scales $n > 0$, $G^n_x$ accepts as input both $z^{n}$ and an upscaled image of $\overline{x}^{n-1}$ to the dimension of $x^n$, denoted by $\uparrow\overline{x}^{n-1}$. A residual generation is then used:
\begin{align}
\overline{x}^{n} = G^{n}_x(z^n + \uparrow\overline{x}^{n-1})  + \uparrow\overline{x}^{n-1}\,. \label{eq:uncon2_res}
\end{align}
This way, the network $ G^{n}_x$ adds the missing details of $\uparrow\overline{x}^{n-1}$ that are specific to scale $n$.

\noindent\textbf{Transformations}\quad
\label{sec:transformations}
To enhance the ability of our model to represent $p^n_{x}$ and inspired by recent work~\cite{zhao2020differentiable,karras2020training} on few shot generation, we apply a fixed set of \textbf{$\mathcal{M}$} differentiable transformations, $T_1, \dots, T_M$, to all real and generated images given as input to $D_x^n$. 
These transformations are chosen, such that, on one hand, they enrich the sample space captured by the current model, thus enabling it to more faithfully capture $p^n_{x}$, while on the other hand, they do not produce samples outside the class of $x$.

We use the set of transformations which result from applying the following transformations sequentially:  (1) horizontal flip, (2) translation: the image is shifted by a ratio of $0.15$ in $x$ axis, y axis or both, (3) $90^{\circ}$ rotations $\{\mathcal{R}_0, \mathcal{R}_{90}, \mathcal{R}_{180}, \mathcal{R}_{270}\}$, and  (4) color transformation of RGB to gray-scale. 
For grayscale image datasets, we do not use color transformations (4). 
Due to memory constraints, we use a subset of $M=54$ ($42$ for grayscale image datasets) transformations from this group, where the same set of transformations are used for all datasets and experiments. The exact set of transformations is given in the supplementary.
We avoid undesirable effects at image borders, by first padding the image with reflection. 
As a pre-processing step, before applying the transformations, we apply histogram equalization to all the training images in RGB color space.

\noindent\textbf{Training objectives}\quad
\label{sec:losses}
$T_1, \dots, T_M$ are used in training $G_x^n$ and $D_x^n$. For each scale, we transform $x^n$ into the set $\{T_1(x^n), \dots, T_M(x^n)\}$ of M transformed real images. Similarly, each generated sample $\overline{x}^n$ is transformed into  $\{T_1(\overline{x}^n), \dots, T_M(\overline{x}^n)\}$. Additionally, the task of recognizing the underlying transformation enriches the output space of $D_x^n$, in a way that is suitable for anomaly detection. 

$D^n_x$ is a fully convolutional Markovian discriminator~\cite{CycleGAN2017, disc2} of the same architecture as $G^n_x$ except that the number of output channels in the last convolutional layer is $\mathcal{M} + 1$. $D^n_x$ is trained in a discriminative fashion to classify all the patches of $T_i(x^n)$ as $i$. In addition, it is optimized to classify all the patches of $T_i(\overline{x}^n)$ to the ``fake'' class $0$. $G^n_x$ tries to fool $D^n_x$ by producing $\overline{x}^n$ such that all the patches of $T_i(\overline{x}^n)$ are classified as $i$.
For each $T_i$, $D^n_x$ produces $M+1$ maps of size $H \times H$. 
Applying softmax, along the $M+1$ dimensions, produces pseudo-probabilities for patch $p$ of the input to belong to one of the $M+1$ classes. We use the following adversarial loss terms:
\begin{align}
\mathcal{L}^D_{adv}(D^n_x) = \sum_{i=1}^M\sum_{p \in H \times H}& \mathcal{L}_{CE}(D^n_x(T_i(x^n))_p, i) \\ -&  \mathcal{L}_{CE}(D^n_x(T_i(\overline{x}^n))_p, 0) \\
\mathcal{L}^G_{adv}(G^n_x) = \sum_{i=1}^M\sum_{p \in H \times H}& \mathcal{L}_{CE}(D^n_x(T_i(\overline{x}^n))_p, i) \\ 
\mathcal{L}_{adv}(D^n_x, G^n_x) = \mathcal{L}^G_{adv}(G^n_x) & - \mathcal{L}^D_{adv}(D^n_x) \label{eq:adv_single}
\end{align}
where $D^n_x(T_i(x^n))_p$ (similarly $D^n_x(T_i(\overline{x}^n))_p$) denotes the softmax probability vector of size $M+1$ for point $p$ in the $H \times H$ map produced by $D^n_x$ and $\mathcal{L}_{CE}$ denotes the cross entropy loss. Recent literature~\cite{golan2018deep} has shown the potential of training a classifier on transformation detection for 
anomaly detection. We utilize a similar discriminative objective, but in the context of a multi-scale hierarchical generative model. As shown in Sec.~\ref{sec:experiments}, the use of this hierarchical generative mode significantly improves results.

In addition to adversarial training, a reconstruction loss is used. 
For $n=0$, $G^0_x$ attempts to reconstruct $x_0$ given a fixed random noise $z^*$ while for $n>0$, $G^n_x$ attempts to reconstruct $x^n$ given an upsampled version of $\overline{x}^{n-1}$ that is obtained, recursively, based on $z^*$, without further randomness:
\begin{align}
\overline{\overline{x}}^{0} &= G^{0}_x(z^{*}) \\
\overline{\overline{x}}^{n} & = G^{n}_x(\uparrow\overline{\overline{x}}^{n-1}) \text{, for $n>0$}\\
\mathcal{L}_{recon_0}(G^0_x) &= ||\overline{\overline{x}}^{0} - x^0||_2 \\
\mathcal{L}_{recon_n}(G^n_x) &= ||\overline{\overline{x}}^{n} - x^n||_2 \text{, for $n>0$}
\end{align}

The reconstruction loss is used to control $\sigma^n$, the standard deviation of the Gaussian noise used at each scale, $z^n$, which indicates the level of detail required at each scale. In particular, $\sigma^n = ||\uparrow\overline{\overline{x}}^{n-1} - x^n||_2$. Without it (using uniform randomness), $G_x^n$ can tint input images.
The overall loss used at scale $n$, for some hyperparameter $\alpha>0$, is then:
\begin{align}
\label{eq:final_gan_loss}
\min_{G^n_x} \max_{D^n_x} \mathcal{L}_{adv}(D^n_x, G^n_x) + \alpha \mathcal{L}_{recon_n}(G^n_x)
\end{align}

\noindent\textbf{Anomaly score}\quad
We cast the problem of anomaly detection as determining if the patches of a given test image, $x_{test}$, and their transformations, are real. That is, for a given test image $x_{test}$, and for each scale $n$, let $T_1(x^n_{test}), \dots, T_M(x^n_{test})$ be the result of applying M transformations on $x^n_{test}$, the down-sampled version of $x_{test}$ at scale $n$. $D^n_x(T_i(x^n_{test}))$ can be viewed as a set of $H\times H$ vectors of size $M+1$, where each vector corresponds to a given patch of the input. By removing the 0'th element (the ``fake" class) and applying softmax, the $i$'th element of this vector denotes the probability that this patch is of class $i$ (the $i$'th transformation). Since $D$ is trained on data from the normal class, the higher this value is, across all patches and scales, and for all transformations $T_i$, the higher our confidence in this sample being non-anomalous. Our anomaly score is, therefore, given as:
\begin{align}
C_{\{x\}}(x_{test}) = \sum_{n=0}^{N} \sum_{i=1}^M \sum_{p \in H \times H} \left[ D^{n*}_x(T_i(x_{test}^n))_p \right]_i \label{eq:score_single}
\end{align}
where $D^{n*}_x$ outputs the last M maps of $D^{n}_x$, i.e., it outputs $H \times H$ vectors of size M, without the 0'th ``fake" element. $D^{n*}_x(T_i(x_{test}^n))_p$ denotes the softmax probability vector of the $p$'the patch, and the index $i$ is indexing this vector, to provide the pseudo-probability of the patch belonging to class (transformation) $i$. The lower $C_{\{x\}}(x_{test})$, the more anomalous $x_{test}$ is.

\subsection{From one shot to few shot anomaly detection}
\label{sec:multi}

Moving to the few shot setting, we are now equipped with a subset $X_k = \{x_1, \dots, x_k \}$ of $X$. To maximize our sample space, we would like to capture the inter-class variability offered by each of the $x_i$'s. 
We, therefore, devise a model that captures the multi-scale patch distribution of each $x_i$. One possibility is to train a model for each $x_i$ separately, and combine the scores of each model.  However, this may be computationally expensive and time consuming.  Instead, we use a single generative model that is conditioned on $i$. 

\noindent\textbf{Conditional generation}\quad
To condition the generator on each training sample $x_i$, we concatenate a single channel, whose values are $i$ everywhere, to the input $z$ of the generator, obtaining an input, which we denote as $cat(z,i)$. 

Let $x$ be a tensor of dimensions $C\times H \times W$ (channels, height and width). We denote by $cat(x, i)$ the tensor of dimensions $(C+1)\times H \times W$, where the last channel equals $i$ in all $H\times W$ positions. 
Generalizing our one-image generator, the generator trained on the set $X_k$ is denoted by $G^{n}_{X_k}$ and is defined as follows:
\begin{align}
\overline{x}_i^{0} &= G^{0}_{X_k}(cat(z^{0}, i)) \label{eq:uncon1_i} \\
\overline{x}_i^{n} &= G^{n}_{X_k}(cat(z^n + \uparrow\overline{x}_i^{n-1}, i))  + \uparrow\overline{x}_i^{n-1} \label{eq:uncon2_res_i}
\end{align}




\noindent\textbf{Training objectives}\quad
Our losses extend those used in Sec.~\ref{sec:losses} to the few shot case. 
Let $D^n_{X_k}$ be the discriminator at scale $n$ trained on $X_k$. We define $\mathcal{L}^i_{adv}(D^n_{X_k}, G^n_{X_k})$ to be $\mathcal{L}_{adv}(D^n_{X_k}, G^n_{X_k})$ (Eq.~\ref{eq:adv_single}) applied with input $x^n_i$ instead of $x^n$. $\mathcal{L}^{multi}_{adv}(D^n_{X_k}, G^n_{X_k})$ is defined to be the sum over $\mathcal{L}^i_{adv}(D^n_{X_k}, G^n_{X_k})$ for all $i$. 
The reconstruction loss for each sample $x_i$, at scale $n$, now becomes: 
\begin{align}
\overline{\overline{x}}_i^{0} &= G^{0}_{X_k}(cat(z^{*}, i)) \\
\overline{\overline{x}}_i^{n} & = G^{n}_{X_k}(cat(\uparrow\overline{\overline{x}}_i^{n-1},i)) \text{, for $n>0$}\\
\mathcal{L}^i_{recon_0}(G^0_{X_k}) &= ||\overline{\overline{x}}_i^{0} - x_i^0||_2 \\
\mathcal{L}^i_{recon_n}(G^n_{X_k}) &= ||\overline{\overline{x}}_i^{n} - x_i^n||_2 \text{, for $n>0$} \nonumber
\end{align}
The full reconstruction loss $\mathcal{L}^{multi}_{recon_n}(G^n_{X_k})$ is the sum over $\mathcal{L}^i_{recon_n}(G^n_{X_k})$ for all $i$. Note that the generator, in the multi-shot case, learns to generate from multiple image distributions $i$. In other words, we have a set of $k$ derived generators per scale $n$, each of which is given by $G^{n}_{X_k}(cat(\cdot, i))$. Similar to the one-shot case, the overall loss at scale $n$ is:
\begin{align}
\label{eq:final_gan_loss_muilti}
\min_{G^n_{X_k}} \max_{D^n_{X_k}} \mathcal{L}^{multi}_{adv}(D^n_{X_k}, G^n_{X_k}) + \alpha \mathcal{L}^{multi}_{recon_n}(G^n_{X_k})
\end{align}
The architecture of $D^n_{X_k}$ is not modified from the one-shot setting. The difference is that now, $D^n_{X_k}$ is trained on more samples $X_k$ as well as generated samples for each sample $i$, and so captures a richer distribution. The anomaly score used in the $k>1$ case, denoted $C_{X_k}(x_{test})$, is the same anomaly score of Eq.~\ref{eq:score_single} with  $D^n_{X_k}$ instead of $D^n_{x}$.

\subsection{Defect Detection}
\label{sec:defect}
We also consider the task of ``defect detection''~\cite{koch2015review}, which is a localized variant of anomaly detection. In this variant, normal samples are visually similar and anomalous samples contain subtle local changes. 
Since our method models the multi-scale patch distribution of each training image, it can be readily used to localize areas where anomalous features occur. 
For anomalous samples, only a small number of patches are anomalous, and the rest are similar to patches in normal samples. 
Therefore, instead of averaging over all patches, we average over the $5\%$ of patches with the lowest anomaly score.
Specifically, Eq.~\ref{eq:score_single} is modified to:
\begin{align}
C_{\{x\}}(x_{test}) &= \sum_{n=0}^{N}\sum_{i=1}^M \sum_{p \in f^i_n(H\times H)} \left[ D^{n*}_x(T_i(x_{test}^n))_p \right]_i \label{eq:score_single_defect} \nonumber
\end{align}
where $f^i_n(H\times H)$ denotes $5\%$ of patch indices with lowest anomaly score for scale $n$ and transformation $i$. The few shot score is defined with $D^n_{X_k}$ instead of $D^n_{x}$. The effect of using a different percentage is given in Sec.~\ref{sec:ablation}. 


\subsection{Architecture and training details} 
\label{sec:architecture}
The generator $G^n_{X_k}$ and the discriminator $D^n_{X_k}$ each consist of five convolutional blocks. A convolutional  block consists of: (i) a $3\times 3$ convolutional layer and is padded such that it maintains the spatial resolution of the input, (ii) Batch normalization layer and (iii) LeakyReLU activation with a slope of $0.2$. The application of five such blocks results in a fixed effective receptive field of $11\times11$ for generator $G^n_{X_k}$ and discriminator $D^n_{X_k}$ at each scale $n$. 
For the last convolutional block, we do not use batch normalization and a $tanh$ is used instead of a LeakyReLU. An adam optimizer learning rate of $0.0005$ and parameters $\beta_1 = 0.5$ and $\beta_2 = 0.999$, are used. 
In Eq.~\ref{eq:final_gan_loss} and Eq.~\ref{eq:final_gan_loss_muilti}, $\alpha = 100.0$ for all experiments.  
$r$ is chosen to be $0.75$ and $N$ is chosen such that the maximal resolution at the finest scale $N$ is $64 \times 64$. 

\begin{figure*}
\centering
\begin{tabular}{c}
\includegraphics[width=0.95\linewidth]{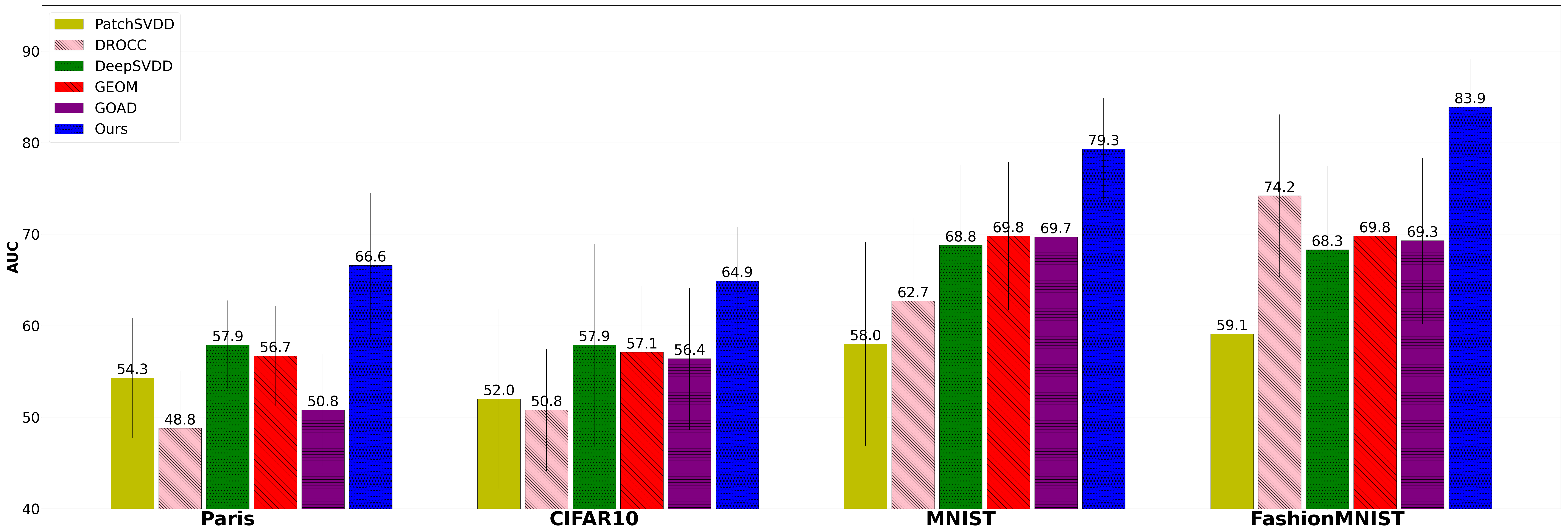} \vspace{-0.2cm} \\
(a) \\
\includegraphics[width=0.95\linewidth]{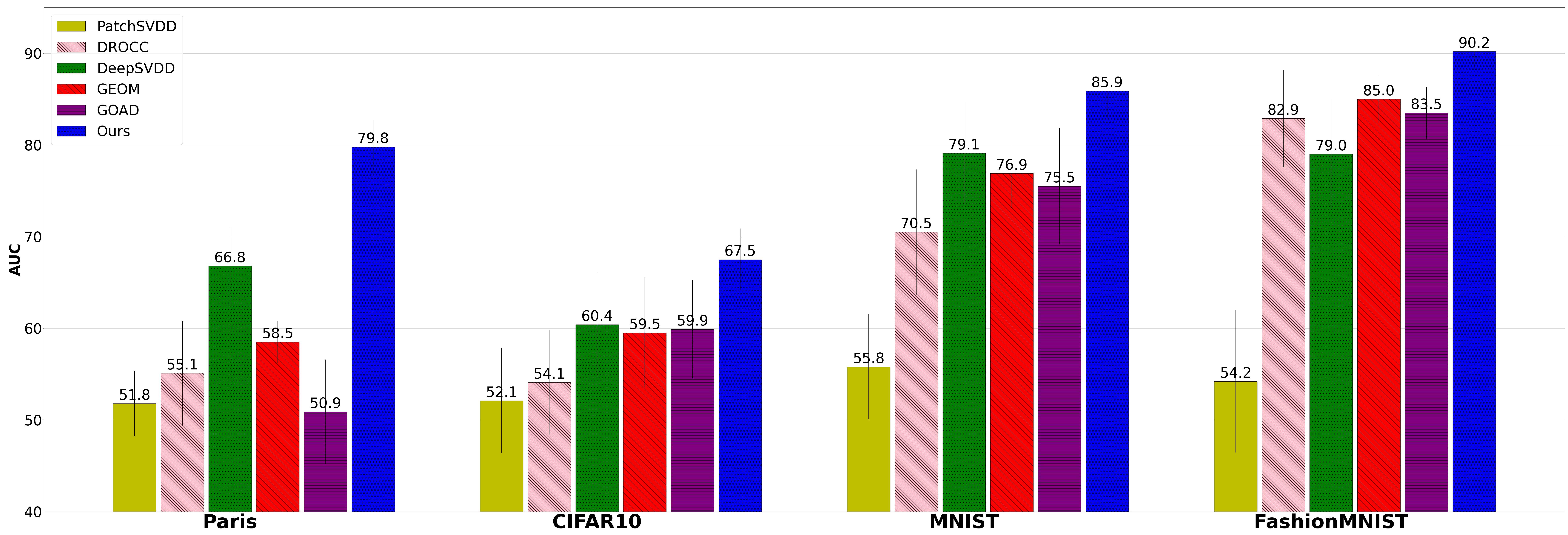} 
\vspace{-0.2cm} \\
(b) \\
\includegraphics[width=0.95\linewidth]{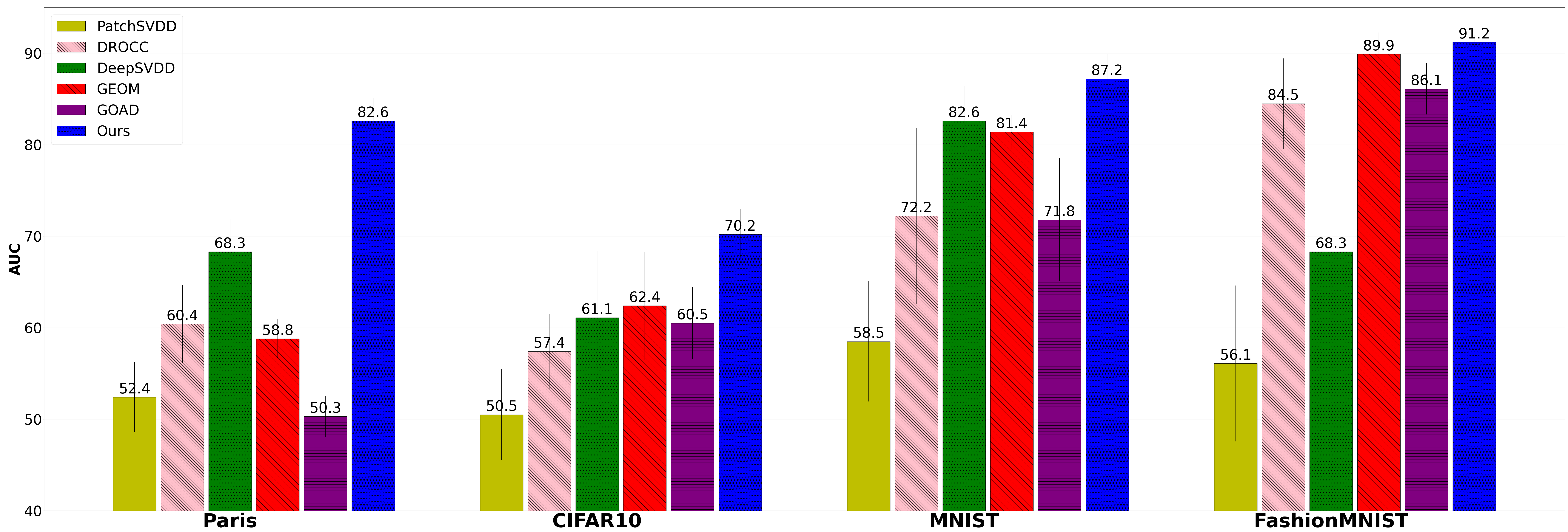} 
\vspace{-0.2cm} \\
(c) \\
\end{tabular}
\caption{Average AUC  (STD indicated with a vertical line) for one and few shot anomaly detection experiments on Paris, CIFAR10, FashionMNIST and MNIST datasets. 
(a): One Shot ($k=1$), (b): Five Shot ($k=5$) and (c): Ten Shot ($k=10$). } 
\label{fig:results}
\vspace{-0.3cm}
\end{figure*}

\begin{figure*}
\centering
\begin{tabular}{c}
\includegraphics[width=0.95\linewidth]{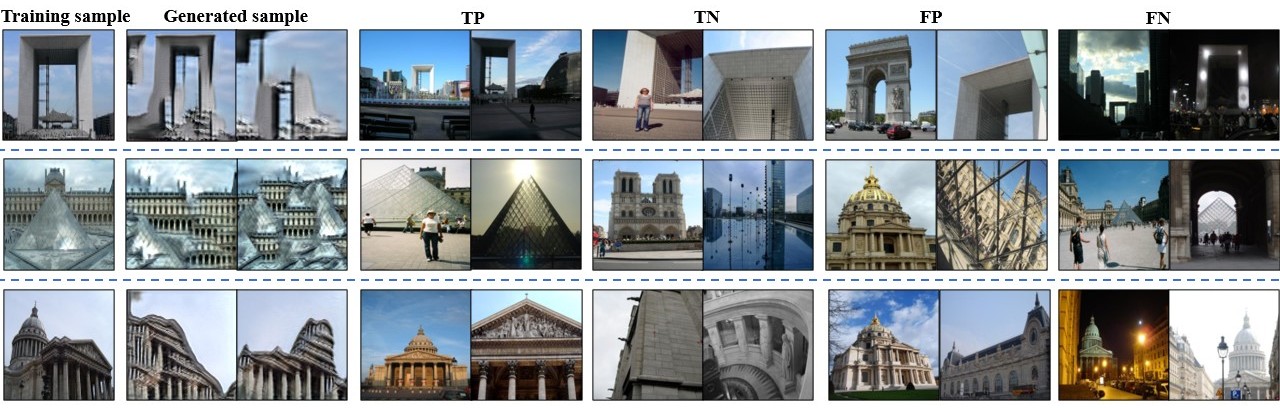}
\\
\end{tabular}
\caption{Illustration of classification decisions made by a single-shot model trained on the Paris dataset. The first column is the training sample, then are random samples generated by the trained generative model. The other columns present samples from the test set of the Paris dataset that are either true positive (\textbf{TP}), true negative (\textbf{TN}), false positive (\textbf{FP}) or false negative (\textbf{TN}).} 
\label{fig:visualization_2}
\vspace{-0.3cm}
\end{figure*}

\begin{figure}
\centering
\begin{tabular}{c}
\includegraphics[width=0.95\linewidth]{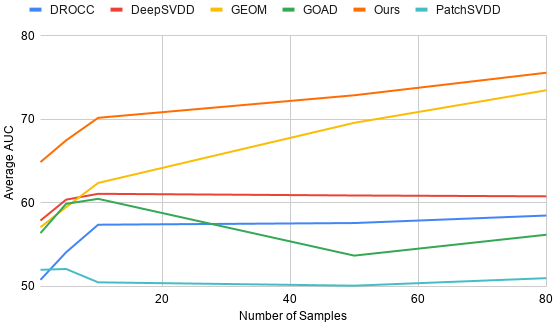} \\
\end{tabular}
\caption{The effect of increasing the number of samples for our method and baselines on CIFAR10. }
\label{fig:increasing_samples}
\vspace{-0.3cm}
\end{figure}

\begin{figure*}
\centering
\begin{tabular}{c}
\includegraphics[width=0.95\linewidth]{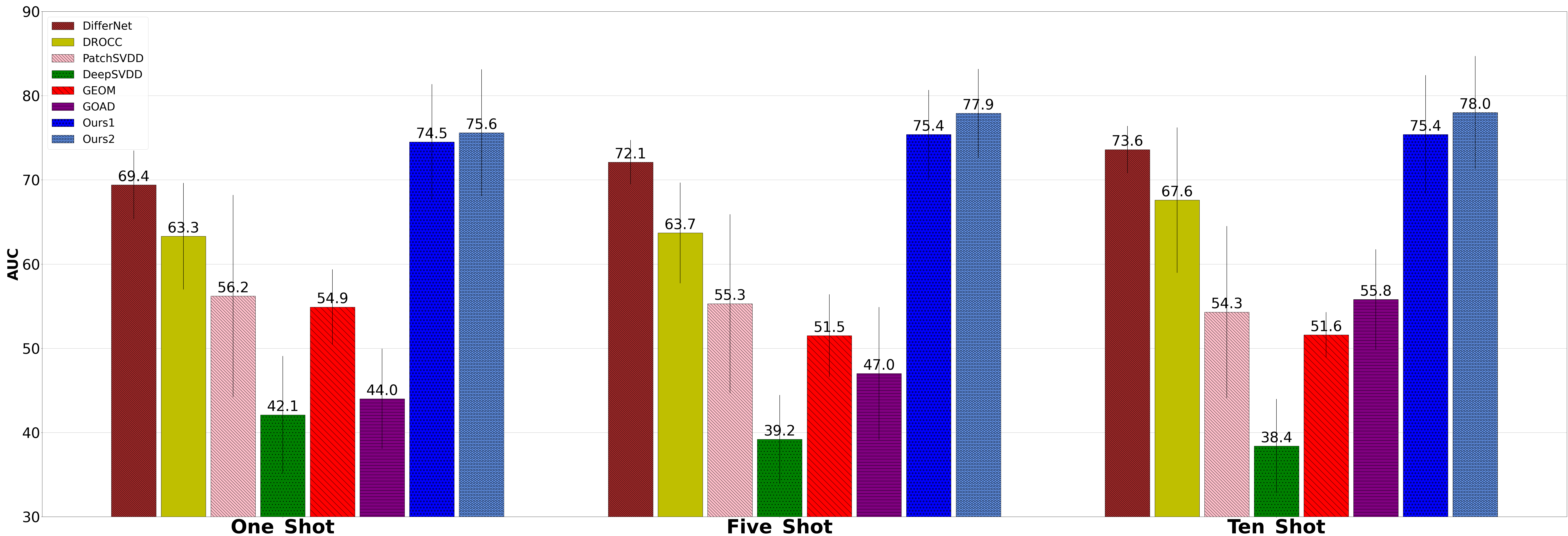} \\
\end{tabular}
\caption{ 
Average AUC (STD indicated with a vertical line) for defect detection on MVTec, for the \textbf{One Shot}, \textbf{Five Shot} and \textbf{Ten Shot} settings. For \textbf{Ours1}, the transformations of anomaly detection (Sec.~\ref{sec:transformations}) are used, while for \textbf{Ours2}, only rotations are used (as in DifferNet). 
} 
\label{fig:defect}
\vspace{-0.3cm}
\end{figure*}

\begin{figure}
\centering
\begin{tabular}{c c c}
\includegraphics[width=0.27\linewidth]{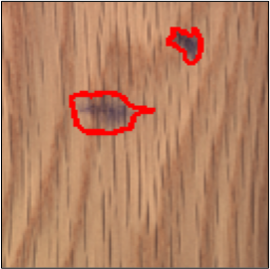} &
\includegraphics[width=0.27\linewidth]{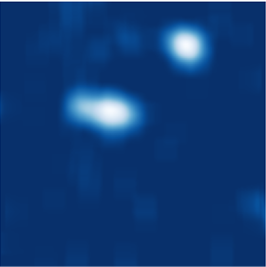} & 
\includegraphics[width=0.27\linewidth]{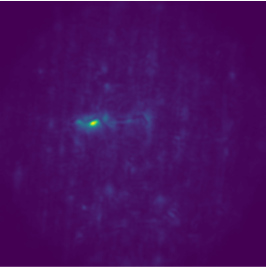} \\
\includegraphics[width=0.27\linewidth]{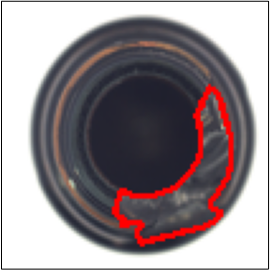} &
\includegraphics[width=0.27\linewidth]{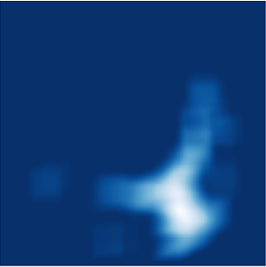} & 
\includegraphics[width=0.27\linewidth]{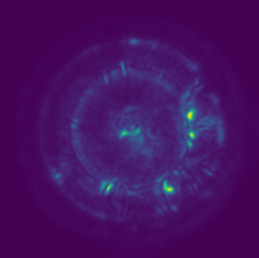} \\
\includegraphics[width=0.27\linewidth]{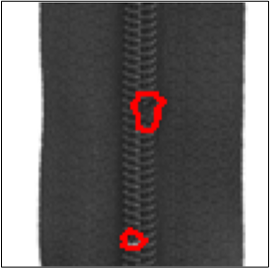} &
\includegraphics[width=0.27\linewidth]{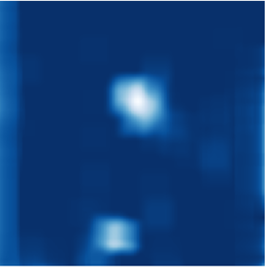} & 
\includegraphics[width=0.27\linewidth]{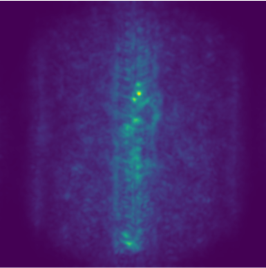} \\
(a) & (b) & (c)
\end{tabular}
\caption{Localization of defects in MVTec test images for one-shot defect detection. (a) The original images, in which the anomaly region is delineated in red. (b-c) The localization provided by (b) our method and (c) DifferNet.} 
\label{fig:visualization_defect}
\end{figure}

\begin{figure}
\centering
\includegraphics[width=0.95\linewidth]{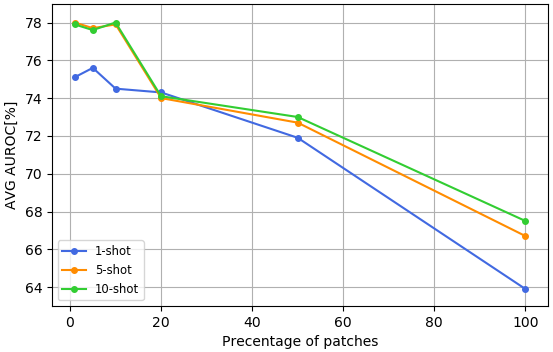} \\
\caption{Effect of percentage of patches used for defect detection.}
\label{fig:defect_sensitivity}
\vspace{-0.3cm}
\end{figure}

\section{Experiments}
\label{sec:experiments}
In all experiments, images are resized to a resolution of $64\times 64$. For ten independent trials and for each class label separately, we train our model on $k$ different images randomly selected from the normal class of images $X$. The same set of images are used for our method and for baselines. At test time, for each trial, using all test images, we measure the Area Under the Curve (AUC). We report the mean AUC and standard deviation values in the main text, and the detailed results per class in the supplementary.

\subsection{Anomaly detection} 
We evaluate our method on four datasets: Paris~\cite{philbin2008lost},  CIFAR10~\cite{krizhevsky2009learning}, FashionMNIST~\cite{xiao2017online} and MNIST~\cite{mnist}. The \textbf{Paris} dataset consists of 6,392 high resolution ($1024 \times 768$) images obtained from Flicker and divided into $11$ landmarks.  \textbf{CIFAR10} consists of $60,000$ $32 \times 32$ color images in $10$ classes, split to $50,000/10,000$ between train and test. \textbf{MNIST} and \textbf{FashionMNIST} both consist of $70,000$  $28 \times 28$ grayscale images of digits or fashion products, respectively, split to $60,000/10,000$ between train and test. There are $10$ categories in each. 
Following~\cite{schlegl2017unsupervised, bergman2020classification, ruff2018deep}, we consider as normal images, all images from a given class, and as anomalous images, the images of all other classes. 

We compare our method to five recent baseline methods. The first method is GEOM~\cite{golan2018deep} which applies different transformations to normal images and trains a classifier to classify the
transformation applied. The second work is GOAD~\cite{bergman2020classification}, which modifies the anomaly score used by GEOM.  
We also consider DROCC~\cite{goyal2020drocc}, that trains a classifier to distinguish the training samples from their perturbations generated adversarially. Lastly, we consider the method of DeepSVDD~\cite{ruff2018deep}, which uses a similar objective to that of
classic SVDD~\cite{scholkopf1999support} together with features of a deep network, and PatchSVDD~\cite{yi2020patch} which extends DeepSVDD to a patch-based method using self-supervised learning. 

\noindent\textbf{One shot anomaly detection}\quad
Fig.~\ref{fig:results}(a) shows the result of our method in comparison to baselines.
Our model outperforms all baselines for all datasets. It scores best for FashionMNIST and MNIST, where a single image captures significant inter-class variability. PatchSVDD, which incorporates a patch-based formulation, scores significantly lower, indicating that, using patches at different scales, together with a our generative and discriminative formulation, significantly improves results. DeepSVDD scores better then PatchSVDD, indicating that deep network features are important.  DROCC, GEOAM and GOAD, which use a discriminative objective, are also inferior to our method, indicating that our patch based generative model is important.   

Qualitative results for the Paris dataset on the one-shot setting are presented in Fig.~\ref{fig:visualization_2}, for typical samples from the ``Defense", ``Louvre" and ``Pantheon" classes. For each class we train the model on random training image and show: (1) the training image, (2) two randomly generated samples, (3) true positive, true negative, false positive and false negative predictions. As can be seen, the generated images capture the appearance and texture of the class, but sometimes alter the structure.
From the true positives, we observe that the model correctly classifies non trivial images that differ from the training image in color, orientation, zoom and even partial occlusions. The false positive (anomalous classified as normal) are often similar to the training sample. The false negatives are often images with significant occlusions or images where the landmark is very small and distant.

\noindent\textbf{Few shot anomaly detection}\quad
We evaluate our method both on $k=5$ and $k=10$ images. Fig.~\ref{fig:results}(b-c) shows the result of our method in comparison to baselines. Our model outperforms the baselines on all datasets, on both the five-shot and ten-shot settings. Increasing the number of samples boosts performance, in particular for the Paris dataset. where using five samples, instead of one, increased the AUC from 66.6\%  to 79.8\%, indicating that multiple images are required to handle inter-class variability. On this dataset, our gap to baselines is also largest, indicating that our method better utilizes the additional variability provided by the additional samples.
In Fig.~\ref{fig:increasing_samples} we consider the effect of increasing the number of samples used for training our method and baselines, on CIFAR10. Our method consistently improves with added number of samples and outperforms baselines on both the 50-shot and 80-shot settings. However, the gap from GEOM decreases as $k$ increases, indicating that when many samples are available, the improvement gained from generating samples at multiple scales diminishes. 

\subsection{Defect detection} 
\label{sec:exp_defect}
For the task of defect detection, we evaluate our method on the MVTec dataset~\cite{bergmann2019mvtec}, which contains $5354$ high-resolution color images of 10 object and 15 texture categories. These are split to $3629/1725$ images between train and test.  The number of training samples per category ranges from $60$ to $320$. A total of 70 defect types, such as little cracks, deformations, discolorizations and scratches occur in test images. The anomalies may differ in size, shape and structure.
For each class, we consider the normal class to be all the defect-free images in this class, and the anomalous images to be all the defective images from the same class. 

We compare our method to DifferNet~\cite{rudolph2021same}  and PatchSVDD~\cite{yi2020patch} that excel in defect detection.
We also compare to anomaly detection methods described above.
We follow the same evaluation protocol for one-shot and few shot anomaly detection. We consider the standard set of transformations as described in Sec.~\ref{sec:transformations} (Ours1 in Fig.~\ref{fig:defect}). For a fair comparison with DifferNet we also consider only the four rotation transformations (group (4) in Sec.~\ref{sec:transformations}) as applied in DifferNet (Ours2 in Fig.~\ref{fig:defect}). 
As can be seen in Fig.~\ref{fig:defect}, our method outperforms all baselines. 


Fig.~\ref{fig:visualization_defect} illustrates the localization of the defects for one-shot defect detection, for both our method and for DifferNet. Three random test samples from MVTec are shown. We visualize the defects at the final scale. Following the notation of Sec.~\ref{sec:defect}, our visualization map is defined as $\sum_{i=1}^M T^{-1}_i \left[ D^{N*}_x(T_i(x_{test}^N))_{(:)} \right]_i$, 
where $\left[ D^{N*}_x(T_i(x_{test}^N))_{(:)} \right]_i $ is an $H \times H$ map indicating how real $T_i(x_{test}^N)$'s patches are. We consider only the patch indices in $f^N_i(H \times H)$, and set the other patches to $0$.  
For DifferNet, we use the visualization procedure provided in their work.
Our method accurately captures the defect regions whereas DifferNet only captures smaller regions of the defect area. 

\begin{table}
    \centering
    \begin{tabular}{ccccccc}
    \toprule
    v. & G & T & H & 1-shot & 5-shot\\
    \midrule
    Full & Yes & Yes & Yes & \textbf{64.9} & \textbf{67.5} \\ 
    \midrule
    (a) & No & Yes & Yes & 60.7 & 64.9\\
    (b) & Yes & No & Yes & 59.1 & 60.0 \\ 
    (c) & Yes & Augment & Yes & 59.7 & 63.4 \\ 
    (d)&  Yes & Yes & No (s=100) & 57.6 & 57.8\\
    (e) & Yes & Yes & No (s=20) & 57.3 & 63.8  \\
    (f) & No & No & No & 47.7 & 48.0\\
    (g) & \multicolumn{3}{c}{Generation followed by GEOM} & 58.8 & 63.7 \\
    \bottomrule
    \\ 
    \end{tabular}
    \caption{Average AUC for CIFAR10, with or without each of the main components of our method: (G) A generator model, (T) Employing transformations, (H) Hierarchy of patches.  See more details in  Sec.~\ref{sec:ablation}.  
    }
    \label{tab:ablation}
    \vspace{-0.3cm}
\end{table}

\subsection{Ablation analysis}
\label{sec:ablation}
Our method relies on three main components: (1) a generative model, (2) its hierarchical multi-scale nature, and (3) a transformation-discriminating component. We assess the contribution of these components separately, running ablation experiments on CIFAR10 for both one-shot and five-shot settings.
A first variant (variant (a), or v.(a) for short) does not have a \textbf{generative component (G)}. Instead, we only use $D^x_n$, and remove the fake class $0$. $D^x_n$ is trained to classify between real images at this scale and their transformations (no fake images are used). The anomaly score remains the same. The second variant (v.(b)) does not employ \textbf{transformations discriminatively (T)}. $D^x_n$ is trained to distinguish between real and fake images at scale $n$, and not between transformations of images.  This is equivalent to setting $M=1$ and using $T_1$ as identity. As another alternative (v.(c)), we apply $T_1, \dots, T_M$  as augmentations (uniformly at random) before being fed to $D^x_n$. 
The next variants consider a single scale of the \textbf{hierarchy (H)}, by setting $N=0$.
We use the same $11 \times 11$ receptive field, and downscale the image to either $100 \times 100$, where small patches are considered, or to $20 \times 20$, where large patches  are considered. This is indicated by $s=100$ (v.(d)) and $s=20$ (v.(e)).
We also consider a simple baseline where no component is used. The anomaly score is the MSE between the test image and the training image in the one-shot setting, and the average MSE in the five-shot setting (v.(f)). Finally, variant v.(g), trains a GEOM~\cite{golan2018deep} model on $6,000$ samples generated using our generative model. 

The results are reported in Tab.~\ref{tab:ablation}. All three components (G, T and H) are required to achieve best performance, with H (hierarchy of patches) playing a particularly important role. Applying GEOM to images generated by out networks is also not as effective as our method. However, it is more effective than running GEOM in the few shot setting (Fig.~\ref{fig:results}).

We further analyze the effect of the percentage of patches ($f^i_n(H\times H)$) taken for defect detection (See Sec.~\ref{sec:defect}). Fig.~\ref{fig:defect_sensitivity} gives the average AUC for MVTec as a function of percentage of patches. Using $5\%$ of patches is best for the one-shot setting while $10\%$ is best for five and ten shot settings. However, the results are stable, when this ratio remains low.

\section{Conclusions}

We present a multi-scale hierarchical generative model, which incorporates, within the discriminators, the self-supervised task of classifying transformations. While multiclass descriptors, in the supervised case, are common in conditional GANs, e.g., ~\cite{salimans2016improved}, we are not aware of other methods that combine labels from a SSL task. Also, while discriminators play an important role in adversarial learning, most works do not employ them outside of training a generator or for creating a secondary feature matching loss~\cite{dcgan}. Also unique, as far as we can ascertain, is the training of single-image like GANs on multiple images. This is done by adding a conditioning layer that contains the image index.

Our method presents a very sizable gap in performance in comparison to the state of the art methods for the few-shot case. Admittedly, training becomes more involved with the increase in the number of training images, and the method does not scale well to hundreds or thousands of training images, without further modifications. In the case of a small training set, for which our method was designed, it demonstrates one-class classification capabilities that are surprising given the emphasis in the existing literature on modeling the form of the variability between the training samples. 



 \section{Acknowledgments}
This project has received funding from the European Research Council (ERC) under the European
Unions Horizon 2020 research and innovation programme (grant ERC CoG 725974). The contribution of the first author is part of a Master thesis research conducted at Tel Aviv University.
{\small
\bibliographystyle{ieee_fullname}
\bibliography{singananomaly}
}

\end{document}




\title{Supplementary Material for  \\ A Hierarchical Transformation-Discriminating Generative\\ Model for Few Shot Anomaly Detection}

\author {Shelly Sheynin$^{1}$\thanks{Equal contribution} \qquad Sagie Benaim$^{1}$\samethanks \qquad Lior Wolf$^{1,2}$ \\
$^1$\normalsize The School of Computer Science, Tel Aviv University\\
$^2$\normalsize Facebook AI Research
}

\maketitle
\ificcvfinal\thispagestyle{empty}\fi

\section{Transformations}

As discussed in Sec.~3.1 of the main text, due to memory constraints,
we use a subset of $M = 54$ transformations. Let $T_{rgb2gray}$ be the transformation of an image from RGB to grayscale. $T_{flip}^{1}$ is a horizonal flip and $T_{flip}^{0}$ is the identity transformation. $T_{translate_x}^{b}$ is the horizontal translation along the x-axis by 15$\%$ of the image width, to the left ($b=1$) or to the right ($b=-1$). $b=0$ is the identity translation. 
$T_{translate_y}^{c}$ is the vertical translation along the y-axis by 15$\%$ of the image height, upwards ($c=1$) or downwards ($c=-1$). $c=0$ is the identity translation.
$T_{rotate}^{d}$ stands for the rotation by $d$ degrees, where $d \in \{0,90,180,270\}$. 

\noindent\textbf{$T_1, \dots T_{32}$:} \quad 
$T_{flip}^{a} \circ T_{translate_x}^{b} \circ T_{translate_y}^{c} \circ T_{rotate}^{d}$ where  $a \in \{0,1\}$, $b \in \{0,1\}$, $c \in \{0,1\}$ and $d \in \{0,90,180,270\}$.


\noindent\textbf{$T_{33}, \dots T_{38}$:} 
$T_{flip}^{a} \circ T_{translate_x}^{b} \circ T_{translate_y}^{c}$, 
where $a \in \{0,1\}$, $b \in \{-1,1,0\}$ and $c=-1$. 


\noindent\textbf{$T_{39}, \dots T_{42}$:} 
$T_{flip}^{a} \circ T_{translate_x}^{b} \circ T_{translate_y}^{c}$, where $a \in \{0,1\}$, $b = -1$ and $c \in \{0,1\}$.  


\noindent\textbf{$T_{43}, \dots T_{50}$:}
$T_{rgb2gray} \circ T_{flip}^{a} \circ T_{rotate}^{d}$, where $a \in \{0,1\}$ and $d \in \{0,90,180,270\}$.



\noindent\textbf{$T_{51}, T_{52}$:}
$T_{rgb2gray} \circ T_{translate_x}^{b}$ where $b \in \{-1,1\}$.

\noindent\textbf{$T_{53}, T_{54}$:}
 $T_{rgb2gray} \circ T_{translate_y}^{c}$ where $c \in \{-1,1\}$.

\section{Detailed Per-Class Results}

In Sec.~4 of the main text, for the task of anomaly detection and defect detection, we report mean AUC values and mean standard deviation values, over all classes. Detailed per-class results are provided here. 

In particular, full anomaly detection results for the datasets of Paris, CIFAR10, FashionMNIST and MNIST are given in Tab.~\ref{tab:one_shot} (one-shot), Tab.~\ref{tab:five_shot} (five-shot) and Tab.~\ref{tab:ten_shot} (ten-shot). This supplements Fig.~2 of the main text. $50$-shot and $80$-shot results for CIFAR10 are given in Tab.~\ref{tab:50_shot_anomaly}. Together with tables \ref{tab:one_shot}-\ref{tab:ten_shot}, this supplements Fig.~4 of the main text.

Tab.~\ref{tab:one_shot_defect} gives the full defect detection results on MVTec for one-shot, five-shot and ten-shot settings, supplementing Fig.~5 of the main text. 

Tab.~\ref{tab:ablation}, gives the ablation analysis performed on CIFAR10, for both the one-shot and five-shot settings, supplementing Tab.~1  and discussed in Sec.~4.3 of the main text. 

Lastly, Tab.~\ref{tab:defect_detection_ablation_1shot}, shows the effect of using a different percentage of patches for detect detection, supplementing Fig.~7 and discussed in Sec.~4.3 of the main text.

\begin{table*}
\centering

\scalebox{0.9}{%
\begin{tabular}{ccccccc}
\toprule
Class  & PatchSVDD & DROCC  & DeepSVDD & GEOM & GOAD & \textbf{Ours} \\
\toprule
\multicolumn{7}{c} {PARIS } \\
\midrule
Defense & 57.0 $\pm$ 3.5 & 53.2 $\pm$ 8.0 & 50.1 $\pm$5.0 &59.4 $\pm$ 3.1 &  47.8 $\pm$ 5.9 & \textbf{65.6 $\pm$ 9.9} \\ 
Eiffel  &46.2 $\pm$ 6.2 & 53.3 $\pm$7.9 & 45.8 $\pm$ 6.7  &46.9 $\pm$ 6.0 &  54.6 $\pm$ 3.3 & \textbf{57.8  $\pm$4.5} \\
Invalides  &46.0 $\pm$ 8.2 & 52.3 $\pm$ 5.1 & 50.3 $\pm$ 6.4   &56.1 $\pm$ 2.9 &  52.9 $\pm$ 3.8  &  \textbf{71.0 $\pm$ 6.4}\\
Louvre   &  47.3 $\pm$ 5.5 & 57.5 $\pm$ 3.3 & 50.1 $\pm$ 3.0  & 53.7 $\pm$ 4.5 &  52.6 $\pm$ 3.1  & \textbf{61.7  $\pm$ 7.2} \\
Moulinrouge  & 60.4 $\pm$ 10.2 &43.7 $\pm$ 6.4 & 64.6 $\pm$ 2.1  &9.4 $\pm$ 7.6 &   51.6 $\pm$ 5.9   & \textbf{72.8  $\pm$6.8} \\
Museedorsay  & 55.7 $\pm$ 8.0 & 42.3 $\pm$ 3.7 & 85.9 $\pm$ 1.9  & \textbf{85.1 $\pm$ 2.7} & 49.3 $\pm$ 16.8  & 73.1 $\pm$ 10.2 \\
Notredame  & 52.3 $\pm$ 4.8 & 46.9 $\pm$ 4.6 & 58.5 $\pm$ 3.1 &52.2 $\pm$ 5.1  & 49.8 $\pm$ 5.7  & \textbf{66.0  $\pm$ 9.4} \\
Pantheon & 62.8 $\pm$ 3.7  & 44.2 $\pm$ 6.6 &  54.8 $\pm$ 12.0 &58.5 $\pm$ 7.8 &  49.9 $\pm$ 5.6 & \textbf{73.8  $\pm$ 8.8}  \\
Pompidou  &56.7 $\pm$ 10.2 & 47.8 $\pm$ 8.9 & 65.5 $\pm$ 3.6  & 65.3 $\pm$ 8.1 &  49 $\pm$ 7.8  & \textbf{68.3  $\pm$9.4}\\
Sacrecoeur & 55.1 $\pm$ 7.9 & 51.8 $\pm$ 8.4  & 52.1 $\pm$ 4.3  & 48.4 $\pm$ 6.7 &  52 $\pm$ 3.5  & \textbf{61.6  $\pm$ 8.5} \\
Triomphe & 57.5 $\pm$ 3.8 &  44.2 $\pm$ 5.9 & 59.2 $\pm$ 5.4  & 48.9 $\pm$ 5.6 & 49 $\pm$ 5.7  & \textbf{60.8  $\pm$ 5.5}\\
\midrule
Avg   & 54.3 $\pm$ 6.5  & 48.8 $\pm$ 6.2 &  57.9  $\pm$ 4.9 & 56.7 $\pm$ 5.5 &  50.8  $\pm$ 6.1 & \textbf{66.6 $\pm$ 7.9}    \\
\midrule 
\vspace{-0.2cm}
\\

\multicolumn{7}{c} {  {CIFAR10} } \\
\midrule 
Plane &  50.1 $\pm$ 15.8 &    54.9 $\pm$ 9.3&   29.8 $\pm$ 5.5    &  49.5 $\pm$ 11.1  & 59.8 $\pm$ 8.3& \textbf{67.2 $\pm$ 5.8} \\
Car   & 51.4 $\pm$ 6.3  &   35.2 $\pm$ 7.4  &    \textbf{81.0 $\pm$ 13.5}  & 53.3 $\pm$ 5.7     & 58.2$\pm$ 5.8 & 65.6 $\pm$ 5.9  \\
Bird  & 46.5 $\pm$ 8.6 &  \textbf{59.5 $\pm$3.7}   &   50.4 $\pm$ 22.4    &54.7 $\pm$ 6.6   & 53.1 $\pm$ 9.1 & 55.9 $\pm$ 5.7 \\
Cat   &48.9 $\pm$ 6.1  &   52.3 $\pm$ 5.5  &   58.8 $\pm$ 12.7   &53.2 $\pm$ 4.4   &  46.4 $\pm$ 8.2 &  \textbf{58.9 $\pm$ 6.2}\\
Deer  & 46.5 $\pm$ 10.7 &   65.7 $\pm$ 5.9  &   56.4 $\pm$ 10.6   &\textbf{67.3 $\pm$ 6.4}  &55.9 $\pm$10.7 & 67.2 $\pm$ 4.5 \\
Dog   & 54.4 $\pm$ 6.3 &    52.7 $\pm$ 8.1 &  22.8 $\pm$ 2.0    &50.9 $\pm$ 2.7 &53.7 $\pm$ 6.0 & \textbf{63.7 $\pm$ 7.7}  \\
Frog  & 53.4 $\pm$ 17.4 &   53.1 $\pm$ 6.8 &  60.2 $\pm$ 15.9    &60.7 $\pm$ 8.6   & 53.6 $\pm$ 9.9  & \textbf{70.2 $\pm$ 5.1} \\
Horse &  52.7 $\pm$ 5.1  & 43.5 $\pm$ 6.1&    \textbf{78.6 $\pm$ 13.1}   & 56.0 $\pm$ 4.6    &  54.8 $\pm$ 7.6 &  63.8 $\pm$ 5.2\\
Ship  &55.6 $\pm$ 13.5 &    57.3 $\pm$ 9.0 &   70.8 $\pm$ 7.9   & 68.1 $\pm$ 10.4    &67.4 $\pm$ 6.4 & \textbf{71.3 $\pm$ 7.2} \\
Truck & 60.8 $\pm$ 8.1 &     33.6 $\pm$ 5.2&   \textbf{69.8 $\pm$ 6.6}    &57.2 $\pm$ 12.0   & 61.1 $\pm$ 5.5 &65.3 $\pm$ 5.2 \\
\midrule
Avg   &  52.0 $\pm$ 9.8  & 50.8 $\pm$ 6.7  & 57.9  $\pm$11.0 &57.1 $\pm$ 7.3 &56.4$\pm$ 7.8 & \textbf{64.9 $\pm$ 5.9}  \\
\midrule \vspace{-0.2cm} \\
\multicolumn{7}{c} {{MNIST} } \\
\midrule 
0 & 46.6 $\pm$ 19.4 & 63.4 $\pm$ 14.1 & \textbf{78.6 $\pm$ 12.7} &  73.1 $\pm$ 5.9  & 77.2 $\pm$ 9.1& 75.2 $\pm$ 5.8 \\
1   & 82.5 $\pm$ 18.1 & 81.6 $\pm$ 5.6  &  69.8 $\pm$ 7.9 &    \textbf{88.7 $\pm$ 5.0}  & 80.2 $\pm$ 18.3& 79.2 $\pm$ 6.9 \\
2  & 56.0 $\pm$ 6.3 & 43.0 $\pm$ 9.2 &67.0 $\pm$ 7.9   & 60.9 $\pm$ 14.4    &72.5 $\pm$ 4.4 &  \textbf{74.3 $\pm$ 3.4} \\
3   & 63.1 $\pm$ 1.7 & 54.3 $\pm$ 8.7 & 61.8 $\pm$ 29.4  &  77.0 $\pm$ 3.2   &80.7 $\pm$ 6.9 & \textbf{94.3 $\pm$ 4.8} \\
4  & 53.6 $\pm$ 8.4 &59.1 $\pm$ 10.4  &   63.2 $\pm$ 5.1& 66.9 $\pm$ 8.4    & 63.8 $\pm$ 5.9& \textbf{81.6 $\pm$ 7.6} \\
5   &60.2 $\pm$ 6.6  & 61.9 $\pm$ 9.5 &   65.2 $\pm$ 4.0& 72.1 $\pm$ 8.3  &54.5 $\pm$ 12.8 & \textbf{80.3 $\pm$ 7.2} \\
6  & 59.0 $\pm$ 11.7 & 65.5 $\pm$ 6.7 &  78.2 $\pm$ 4.9&  66.2 $\pm$ 20.2  &70.2 $\pm$ 4.2 & \textbf{85.7 $\pm$ 3.4} \\
7 &  49.2 $\pm$ 14.0  & 70.1 $\pm$ 12.0 & 70.2 $\pm$ 3.2  &  69.5 $\pm$ 8.9   & 66.4 $\pm$ 10.3&  \textbf{76.9 $\pm$ 4.0} \\
8  & 53.7 $\pm$ 15.6 & 57.5 $\pm$ 7.4 &  \textbf{72.4 $\pm$ 3.7} &   56.2 $\pm$ 2.1  & 71.7 $\pm$ 4.7 & 71.5 $\pm$ 6.2 \\
9 & 56.3 $\pm$ 8.9 & 70.3 $\pm$ 7.2 &  61.8 $\pm$ 9.0 &   67.6 $\pm$ 4.3  &59.8 $\pm$ 5.1 & \textbf{73.5 $\pm$ 6.4} \\
\midrule
Avg   & 58.0 $\pm$ 11.1 & 62.7 $\pm$ 9.1 & 68.8 $\pm$ 8.8  & 69.8 $\pm$ 8.1 & 69.7 $\pm$ 8.2 & \textbf{79.3 $\pm$ 5.6}\\
\midrule \vspace{-0.2cm} \\
\multicolumn{7}{c} {{FashionMNIST}} \\
\midrule 
T-shirt &  58.5 $\pm$ 5.6&   69.7 $\pm$ 8.1 &  \textbf{83.5 $\pm$ 6.9} &  79.7 $\pm$ 2.9 &  71.8 $\pm$ 14.1  & 77.3 $\pm$ 4.3 \\
Trouser   & 32.0 $\pm$ 18.6 &  95.2 $\pm$ 1.7 & 63.5 $\pm$ 9.2   &  55.5 $\pm$ 4.3 &  76.0 $\pm$3.7   & \textbf{97.2 $\pm$ 1.4} \\
Pullover  & 73.7 $\pm$ 8.7 &  68.0 $\pm$ 8.9 &  66.7 $\pm$ 7.3 &  56.9 $\pm$ 12.1   &  69.1 $\pm$5.6   & \textbf{80.3 $\pm$ 4.2} \\
Dress   & 43.0 $\pm$ 9.8 & 80.9 $\pm$ 6.6   & 63.1 $\pm$ 16.3  &    72.5 $\pm$ 10.5 &   76.9 $\pm$13.1  & \textbf{83.8 $\pm$ 4.0}  \\
Coat  & 73.3 $\pm$ 4.9 &63.5 $\pm$ 15.1  & 63.6 $\pm$ 12.0  &   52.2 $\pm$ 16.1  &   66.2 $\pm$18.8  &  \textbf{79.0 $\pm$ 9.2} \\
Sandals  &  39.1 $\pm$ 26.4 & 74.3 $\pm$ 8.4    &64.9 $\pm$ 9.8   &  78.5 $\pm$ 9.7 & 57.9 $\pm$10.1   & \textbf{85.5 $\pm$ 4.5} \\
Shirt  &  70.2 $\pm$ 2.7 &  64.9 $\pm$ 8.9   &  \textbf{75.1 $\pm$ 6.2}   & 56.1 $\pm$ 5.6   & 72.8 $\pm$3.1    & 69.0 $\pm$ 2.4\\
Sneaker &  58.1 $\pm$ 25.7 & 90.5 $\pm$ 9.1 &  59.1 $\pm$ 12.0 &  92.6 $\pm$ 2.1  &  69.2 $\pm$1.7   &  \textbf{97.9 $\pm$ 0.7} \\
Bag  &  70.2 $\pm$ 2.1 &  53.6 $\pm$ 7.4   &  72.4 $\pm$ 3.3  &     \textbf{92.2 $\pm$ 9.1} &  71.7 $\pm$9.9  & 77.2 $\pm$ 15.4\\
Ankle-Boot & 73.2 $\pm$ 9.2  &  81.9 $\pm$ 14.7   &   71.2 $\pm$ 8.5  &   62.0 $\pm$ 5.8  &   61.6 $\pm$10.6  & \textbf{91.7 $\pm$ 6.1} \\
\midrule
Avg   &59.1 $\pm$ 11.4  & 74.2 $\pm$ 8.9 &  68.3 $\pm$ 9.2 &  69.8 $\pm$ 7.8  & 69.3 $\pm$ 9.1 &  \textbf{83.9 $\pm$ 5.2} \\
\bottomrule \\
\end{tabular}}

\caption{ Average AUC  (with standard deviation) for \textbf{One-Shot} anomaly detection experiments on Paris, CIFAR10, FashionMNIST and MNIST datasets. 
}
\label{tab:one_shot}

\end{table*}

\begin{table*}
\centering

\scalebox{0.9}{%
\begin{tabular}{ccccccc}
\toprule
Class  & PatchSVDD & DROCC  & DeepSVDD & GEOM & GOAD & \textbf{Ours} \\
\toprule
\multicolumn{7}{c} {PARIS } \\
\midrule
Defense & 51.5 $\pm$ 3.3 & \textbf{69.3 $\pm$ 4.5} &  62.1 $\pm$ 3.3  & 59.4 $\pm$ 2.7 & 52.8 $\pm$ 5.1 & 67.8 $\pm$ 3.4 \\
Eiffel  & 51.2 $\pm$ 4.3 & 66.8 $\pm$ 3.5 &  55.4 $\pm$ 2.8 & 44.1 $\pm$ 6.6 & 53.0 $\pm$ 3.0 & \textbf{67.0 $\pm$ 2.7} \\
Invalides  &45.2 $\pm$ 2.1  & 62.9 $\pm$ 6.4 & 66.6 $\pm$ 4.9 & 59.2 $\pm$ 2.0 & 52.2 $\pm$ 4.5 & \textbf{80.8 $\pm$ 2.5}  \\
Louvre   & 41.1 $\pm$ 2.0 & 66.6 $\pm$ 3.3 & 60.4 $\pm$ 4.3  &  53.3 $\pm$ 1.8  & 52.3 $\pm$ 2.7  & \textbf{72.5 $\pm$ 2.8} \\
Moulinrouge  & 59.6 $\pm$ 3.1 & 44.1 $\pm$ 5.4 & 62.4 $\pm$ 5.1 & 49.0 $\pm$ 0.3 & 45.9 $\pm$ 7.3& \textbf{84.5 $\pm$ 2.4} \\
Museedorsay   & 53.9 $\pm$ 2.7 & 46.8 $\pm$ 9.6 &  88.0 $\pm$ 3.3 & 88.7 $\pm$ 3.2  &  43.0 $\pm$ 15.2 & \textbf{89.6 $\pm$1.8} \\
Notredame  & 47.7 $\pm$ 2.8 & 48.7 $\pm$ 6.6 &62.6 $\pm$ 3.5  &   58.4 $\pm$ 1.7   & 48.2 $\pm$5.7  & \textbf{79.7 $\pm$ 4.0}\\
Pantheon &  58.4 $\pm$ 5.2  & 49.2 $\pm$ 6.6 &  74.9 $\pm$ 2.5 &   60.7 $\pm$ 1.8  &  52.3 $\pm$ 2.3 & \textbf{86.1 $\pm$ 2.1} \\
Pompidou  & 58.7 $\pm$ 3.4 & 45.7 $\pm$ 7.4 &  75.6 $\pm$ 3.4 & 70.0 $\pm$ 2.7   & 54.1 $\pm$5.8   & \textbf{90.3 $\pm$ 3.4} \\
Sacrecoeur & 46.9 $\pm$ 7.8 & 58.5 $\pm$ 5.1 & 62.5 $\pm$ 4.3  &  48.1 $\pm$ 0.8   &  54.8 $\pm$ 7.2 & \textbf{81.6 $\pm$ 2.7} \\
Triomphe & 55.5 $\pm$ 2.6 &47.4 $\pm$ 4.4  &  64.2 $\pm$ 9.4 &    52.4 $\pm$ 1.7 & 51.1 $\pm$ 3.8 & \textbf{78.6 $\pm$ 4.8} \\
\midrule
Avg   & 51.8 $\pm$ 3.6 & 55.1 $\pm$ 5.7 & 66.8 $\pm$ 4.2  & 58.5 $\pm$ 2.3 & 50.9 $\pm$ 5.7 & \textbf{79.8 $\pm$ 3.0} \\
\midrule \vspace{-0.2cm}\\
\multicolumn{7}{c} {CIFAR10 } \\
\midrule
Plane & 40.8 $\pm$ 13.8 &  69.0 $\pm$ 4.8& 35.8 $\pm$ 3.1 &62.3 $\pm$ 9.0 & 59.5 $\pm$ 3.0 & \textbf{69.2 $\pm$ 2.8} \\
Car   & 59.5 $\pm$ 4.5 & 39.6 $\pm$ 8.8 &  74.6 $\pm$ 5.3 &65.5 $\pm$ 7.7 & 68.8 $\pm$10.1& \textbf{77.0 $\pm$ 1.8} \\
Bird  & 45.7 $\pm$ 6.0 &  \textbf{60.9 $\pm$ 3.4} & 48.4 $\pm$ 5.2  &52.4 $\pm$ 4.8 & 49.4 $\pm$ 3.4 & 58.4 $\pm$ 2.3 \\
Cat   & 55.6 $\pm$ 3.2 & 56.5 $\pm$ 4.9 & 54.4 $\pm$ 10.7  &54.0 $\pm$ 5.2 & 49.0 $\pm$ 7.0& \textbf{58.7 $\pm$ 4.3} \\
Deer  & 44.5 $\pm$ 5.3 & 57.9 $\pm$ 5.4 &  51.4 $\pm$ 5.8 & 63.6 $\pm$ 7.6& 48.8 $\pm$ 5.5 & \textbf{66.4 $\pm$ 4.3} \\
Dog   & 54.4 $\pm$ 3.0 & 59.4 $\pm$ 6.1 & \textbf{70.4 $\pm$ 6.1}  &  55.5 $\pm$ 3.4 &60.9 $\pm$ 11.5 & 61.8 $\pm$ 3.2 \\
Frog  &  53.7 $\pm$ 5.6& 50.2 $\pm$ 7.7 & 56.0 $\pm$ 5.7 & 58.5 $\pm$ 6.9 &51.5 $\pm$ 2.7 & \textbf{72.6 $\pm$ 4.4} \\
Horse &   55.4 $\pm$ 3.2 & 43.6 $\pm$ 4.3 & \textbf{69.7 $\pm$ 5.9}  &  64.2 $\pm$3.1 &62.0 $\pm$ 4.9 & 68.6 $\pm$ 2.8 \\
Ship  &  48.3 $\pm$ 10.3 & 67.5 $\pm$ 6.7 &  73.4 $\pm$ 4.4 & 75.5 $\pm$7.9  & 74.2 $\pm$ 3.6 & \textbf{80.2 $\pm$ 3.2} \\
Truck & 62.6 $\pm$ 2.2 & 35.9 $\pm$ 5.5 &  70.3 $\pm$ 4.7 &  67.5 $\pm$ 4.0 & \textbf{74.2 $\pm$ 1.7} & 62.1 $\pm$ 4.5\\
\midrule
Avg   & 52.1 $\pm$ 5.7 & 54.1 $\pm$ 5.8 &  60.4 $\pm$ 5.7 & 59.5 $\pm$ 6.0  & 59.9 $\pm$ 5.3  & \textbf{67.5 $\pm$ 3.4}   \\
\midrule \vspace{-0.2cm}\\
\multicolumn{7}{c} {MNIST } \\
\midrule
0 & 76.6 $\pm$ 2.5 &   70.7 $\pm$ 9.0 &  86.8 $\pm$ 3.2  &  71.3 $\pm$ 6.3  &  \textbf{87.4 $\pm$ 8.0}  & 79.5 $\pm$ 3.8 \\
1   & 31.5 $\pm$ 9.9 & 80.6 $\pm$ 7.5 &  89.6 $\pm$ 5.3 & \textbf{96.2 $\pm$ 0.5}   &  89.2 $\pm$ 6.8  & 85.5 $\pm$ 6.7 \\
2  & 73.5 $\pm$ 5.2 & 56.4 $\pm$ 10.2 &  73.4 $\pm$ 5.2 &  78.0 $\pm$ 2.6  &  71.3 $\pm$ 7.3  & \textbf{81.6 $\pm$ 4.2} \\
3   &  71.0 $\pm$ 5.5& 63.4 $\pm$ 5.3 & 77.2 $\pm$ 10.7  &   85.5 $\pm$ 0.7  & 80.9 $\pm$ 4.6    & \textbf{96.6 $\pm$ 1.0} \\
4  &45.0 $\pm$ 5.8  & 69.6 $\pm$ 3.5 & 76.8 $\pm$ 5.8  & 66.4 $\pm$ 5.6   &  70.3 $\pm$ 5.2   & \textbf{84.7 $\pm$ 1.3}  \\
5   & 62.6 $\pm$ 3.0 &69.1 $\pm$ 7.2  &  65.6 $\pm$ 6.1 & 79.0 $\pm$ 8.5  & 70.4 $\pm$ 12.8    &  \textbf{89.3 $\pm$ 2.4} \\
6  & 55.5 $\pm$ 4.3 & 73.9 $\pm$ 7.5 & 80.0 $\pm$ 5.7 &   76.1 $\pm$ 4.6  &  72.6 $\pm$ 3.9   & \textbf{92.4 $\pm$ 0.9} \\
7 &   35.2 $\pm$ 8.3 & 80.4 $\pm$ 7.2 & 81.0 $\pm$ 5.9  &  80.3 $\pm$ 3.8  &   67.1 $\pm$ 5.7  & \textbf{82.0 $\pm$ 3.7} \\
8  &  64.9 $\pm$ 6.5& 64.4 $\pm$ 4.6 & \textbf{82.2 $\pm$ 4.4}  &  70.7 $\pm$ 4.0  &73.4 $\pm$ 5.1    & 79.4 $\pm$ 3.4\\
9 &  42.2 $\pm$ 6.4& 76.7 $\pm$ 6.3 & 79.2 $\pm$ 4.7  &  65.7 $\pm$ 1.9  &  72.5 $\pm$ 3.9   & \textbf{87.5 $\pm$ 3.2} \\
\midrule
Avg   & 55.8 $\pm$ 5.7 & 70.5 $\pm$ 6.8 &  79.1 $\pm$ 5.7 & 76.9 $\pm$3.9 & 75.5 $\pm$ 6.3 & \textbf{85.9$\pm$ 3.1} \\
\midrule \vspace{-0.2cm}\\
\multicolumn{7}{c} {FashionMNIST } \\
\midrule
T-shirt &  52.8 $\pm$ 6.0 &  85.2 $\pm$ 3.1  & 89.6 $\pm$ 2.4  &  \textbf{92.4 $\pm$ 2.7} &  79.8 $\pm$ 2.7 &85.2 $\pm$ 1.7  \\
Trouser   & 42.2 $\pm$ 10.7 & 94.2 $\pm$ 2.2 &  84.8 $\pm$ 7.1 & 74.7 $\pm$ 2.7  & 97.8 $\pm$ 0.5  & \textbf{98.4 $\pm$ 0.5} \\
Pullover  &  64.7 $\pm$ 7.0& 80.5 $\pm$ 3.3 & 72.3 $\pm$ 7.1  & 84.3 $\pm$ 3.6  & \textbf{86.4 $\pm$2.2}  & 85.8 $\pm$ 3.5 \\
Dress   &  41.7 $\pm$ 7.6& 86.3 $\pm$ 4.4  & 77.8 $\pm$ 4.9  &  87.8 $\pm$ 1.0  & 85.1 $\pm$ 2.0   & \textbf{89.1 $\pm$ 2.4} \\
Coat  &  62.8 $\pm$ 6.1  & 81.5 $\pm$ 3.9  &  76.8 $\pm$ 7.0   &  78.4 $\pm$ 2.0  &  83.8 $\pm$ 1.9  & \textbf{88.4 $\pm$ 1.5} \\
Sandals   & 60.1 $\pm$ 8.5&  78.1 $\pm$ 15.0  &  63.8 $\pm$ 8.0  & 83.7 $\pm$ 2.0  &  65.9 $\pm$ 7.6  & \textbf{88.6 $\pm$ 2.1}\\
Shirt &   54.8 $\pm$ 6.6 &  72.0 $\pm$ 3.5  &  \textbf{81.5 $\pm$ 8.0}  &   73.8 $\pm$ 3.7  & 68.0 $\pm$ 3.1   &78.2 $\pm$ 1.7 \\
Sneaker &  53.0 $\pm$ 10.7  & 93.2 $\pm$ 1.6 & 81.6 $\pm$ 7.4  &  94.6 $\pm$ 1.9  &  94.4 $\pm$ 1.0   & \textbf{99.1 $\pm$ 0.3} \\
Bag &  53.4 $\pm$ 5.3  &   67.6 $\pm$ 8.7 & 80.1 $\pm$ 3.1    &   \textbf{96.6 $\pm$ 1.4} &   77.5 $\pm$ 6.0 &92.9 $\pm$ 4.1 \\
Ankle-Boot & 56.4 $\pm$ 9.1 &  90.0 $\pm$ 6.9 & 82.1 $\pm$ 5.4   & 83.7 $\pm$ 4.7  &  \textbf{96.6 $\pm$ 1.5}  &96.5 $\pm$ 1.0 \\
\midrule
Avg   &  54.2 $\pm$ 7.8 &   82.9 $\pm$ 5.3  &  79.0 $\pm$ 6.0   &  85.0  $\pm$ 2.6  &  83.5 $\pm$ 2.9 &  \textbf{90.2 $\pm$ 1.9} \\
\bottomrule \\
\end{tabular}}
\caption{Average AUC  (with standard deviation) for \textbf{Five-Shot} anomaly detection experiments on Paris, CIFAR10, FashionMNIST and MNIST datasets. }
\label{tab:five_shot}
\end{table*}

\begin{table*}
\centering
\scalebox{0.9}{%
\begin{tabular}{ccccccc}
\toprule
Class  & PatchSVDD & DROCC  & DeepSVDD & GEOM & GOAD & \textbf{Ours}\\
\toprule
\multicolumn{7}{c} {PARIS } \\
\midrule
Defense & 54.2 $\pm$ 4.1 & \textbf{72.0 $\pm$ 4.5} &  62.7 $\pm$ 2.3&  57.2 $\pm$ 1.6  &   49.5 $\pm$ 3.3 &67.9 $\pm$ 3.2 \\
Eiffel  & 48.1 $\pm$ 4.8 & \textbf{73.6 $\pm$ 3.3}  & 59.4 $\pm$ 1.9  & 47.2 $\pm$ 5.6  & 50.0 $\pm$ 0.0   & 71.2 $\pm$ 3.7 \\
Invalides  & 42.7 $\pm$ 2.8  & 68.0 $\pm$ 4.8 & 67.4 $\pm$ 2.0  & 63.4 $\pm$ 1.0  &  49.9 $\pm$ 0.0  & \textbf{84.9 $\pm$ 1.2}  \\
Louvre   & 39.4 $\pm$ 1.9 & 73.5 $\pm$ 4.3 &   60.6 $\pm$ 3.3 & 53.4 $\pm$ 1.8   &  49.2 $\pm$ 1.6  & \textbf{74.9 $\pm$ 2.5} \\
Moulinrouge  & 59.4 $\pm$ 4.3 & 46.6 $\pm$ 2.5 & 63.6 $\pm$ 3.4  &  51.7 $\pm$ 0.8  &  48.8 $\pm$ 2.2  & \textbf{87.0 $\pm$ 3.1} \\
Museedorsay   & 58.4 $\pm$ 4.0 & 52.3 $\pm$ 3.2 &  89.4 $\pm$ 2.0 & 86.0 $\pm$ 5.1  &    55.5 $\pm$ 6.9 & \textbf{90.7 $\pm$ 2.2} \\
Notredame  & 52.0 $\pm$ 3.2 & 52.5 $\pm$ 4.6 & 65.8 $\pm$ 2.9  & 55.3 $\pm$ 1.0 &  49.9 $\pm$ 3.1  & \textbf{83.0 $\pm$ 2.9} \\
Pantheon &  54.5 $\pm$ 4.9  & 57.2 $\pm$ 6.4 & 75.7 $\pm$ 1.6  & 62.3 $\pm$ 0.8 &   50.3 $\pm$ 0.7  &  \textbf{89.9 $\pm$ 2.1} \\
Pompidou  & 59.9 $\pm$5.6 & 50.3 $\pm$ 4.6 &  77.6 $\pm$ 6.0 & 69.2 $\pm$ 0.8 &   50.2 $\pm$ 2.7  & \textbf{95.4 $\pm$ 1.3} \\
Sacrecoeur & 48.1 $\pm$ 2.4 & 66.6 $\pm$ 4.9 &66.1 $\pm$ 3.4   & 47.1 $\pm$ 4.1 &   51.2 $\pm$ 3.1  & \textbf{84.5 $\pm$ 1.9} \\
Triomphe & 59.5 $\pm$ 4.0 & 51.7 $\pm$ 3.9 &  63.3 $\pm$ 10.3 & 53.4 $\pm$ 0.6 &   49.0 $\pm$ 1.1  & \textbf{79.8 $\pm$ 3.4} \\
\midrule
Avg   & 52.4 $\pm$ 3.8  & 60.4 $\pm$ 4.3 & 68.3  $\pm$ 3.5 & 58.8 $\pm$ 2.1 & 50.3 $\pm$ 2.5 & \textbf{82.6 $\pm$ 2.5} \\
\midrule \vspace{-0.2cm} \\
\multicolumn{7}{c} {CIFAR10 } \\
\midrule
Plane & 40.8 $\pm$ 9.4 & \textbf{71.9 $\pm$ 2.2} & 39.6 $\pm$ 6.3 & 66.7 $\pm$ 8.8 & 61.5$\pm$2.4 & 69.1 $\pm$ 1.6 \\
Car   & 59.9 $\pm$ 3.4 & 42.8 $\pm$ 8.2 & 64.0 $\pm$ 9.9  &   74.3 $\pm$ 2.7 & 68.7$\pm$6.1   & \textbf{80.7 $\pm$ 2.9} \\
Bird  & 44.8 $\pm$ 3.9& \textbf{62.4 $\pm$ 4.4} & 42.4 $\pm$ 11.1  &   54.4 $\pm$ 6.7 &  51.3$\pm$3.2  & 58.5 $\pm$ 2.5 \\
Cat   & 53.8 $\pm$ 3.7 &61.7 $\pm$ 4.3  &  54.3 $\pm$ 7.3 & 52.5 $\pm$ 5.7  &  50.4$\pm$4.8  & \textbf{63.2 $\pm$ 2.8} \\
Deer  & 50.1 $\pm$ 4.9 & 62.0 $\pm$ 3.2 & 50.0 $\pm$ 8.7  &  54.1 $\pm$ 5.8 &   52.1 $\pm$ 7.1 & \textbf{64.2 $\pm$ 2.2} \\
Dog   & 53.3 $\pm$ 4.3 & 61.3 $\pm$ 3.9 & \textbf{81.6 $\pm$ 3.9}   &  60.5 $\pm$ 5.1 &  57.1$\pm$5.7   &65.4 $\pm$ 5.6 \\
Frog  & 50.4 $\pm$ 4.7 &  48.2 $\pm$ 4.6& 58.0 $\pm$ 11.9 & 60.3 $\pm$ 6.8   &  55.3$\pm$2.3   & \textbf{71.9 $\pm$ 3.3} \\
Horse & 53.9 $\pm$ 2.9 & 51.6 $\pm$ 3.1 & \textbf{76.8 $\pm$ 5.4}  & 62.9 $\pm$ 4.5   & 61.7$\pm$3.2    &  73.7 $\pm$ 2.8\\
Ship  &  46.0 $\pm$ 8.5 & 72.6 $\pm$ 3.4 &  71.6 $\pm$ 3.9 &  67.8 $\pm$ 8.7  &  71.3$\pm$2.3   & \textbf{82.9 $\pm$ 0.8} \\
Truck & 52.6 $\pm$ 4.2 & 39.3 $\pm$ 3.5 &  73.4 $\pm$ 4.2 &  70.3 $\pm$ 4.0  &   \textbf{75.2$\pm$2.5} & 72.6 $\pm$ 2.9\\
\midrule
Avg   & 50.5 $\pm$ 5.0 & 57.4 $\pm$ 4.1 & 61.1 $\pm$ 7.3 &  62.4 $\pm$ 5.9 & 60.5 $\pm$ 4.0  & \textbf{70.2 $\pm$ 2.7} \\
\midrule \vspace{-0.2cm} \\
\multicolumn{7}{c} {MNIST } \\
\midrule
0 &  75.0 $\pm$ 4.7 & 80.3 $\pm$ 8.0 & \textbf{91.6 $\pm$ 1.1} & 75.0 $\pm$ 1.0   &  72.6 $\pm$ 6.8 &  80.1 $\pm$ 4.6\\
1   & 59.3 $\pm$ 13.1 & 78.0 $\pm$ 12.5 & 89.0 $\pm$ 5.8  &  \textbf{96.2 $\pm$ 0.4}  & 90.9 $\pm$ 3.2   &88.8 $\pm$ 3.1  \\
2  & 58.6 $\pm$ 5.7 & 58.8 $\pm$ 13.7 & 73.0 $\pm$ 8.8  & 80.1 $\pm$ 2.4   & 68 $\pm$ 5.6  & \textbf{85.2 $\pm$ 4.3} \\
3   & 62.0 $\pm$ 6.1 & 66.9 $\pm$ 7.6 & 82.4 $\pm$ 3.2  & 91.0 $\pm$ 0.5   &73.2 $\pm$ 9.3 &  \textbf{96.3 $\pm$ 0.9}\\
4  & 53.7 $\pm$ 7.8 & 71.2 $\pm$ 9.8 & 85.6 $\pm$ 0.9  & 79.3 $\pm$ 1.1  &69.1 $\pm$ 6.2 & \textbf{89.1 $\pm$ 1.6} \\
5   & 59.8 $\pm$ 5.2 & 63.7 $\pm$ 8.2 &  72.4 $\pm$ 4.0 & 87.2 $\pm$ 0.6  & 62.1 $\pm$ 13.4& \textbf{87.4 $\pm$ 3.3} \\
6  & 53.9 $\pm$ 4.8 &  74.0 $\pm$ 14.3  & 88.2 $\pm$ 2.5   & 83.6 $\pm$ 2.8    &73.9 $\pm$ 4.6 & \textbf{92.2 $\pm$ 1.6} \\
7 &  50.4 $\pm$ 6.5  & 77.1 $\pm$ 10.8 & 80.0 $\pm$ 7.5  &78.4 $\pm$ 0.7   & 63 $\pm$ 6.5& \textbf{84.2 $\pm$ 4.2} \\
8  & 61.5 $\pm$ 4.8 & 69.1 $\pm$ 4.8 & \textbf{81.0 $\pm$ 0.9}  & 64.7 $\pm$ 4.0 &77.8 $\pm$ 5.4 &78.2 $\pm$ 2.3 \\
9 & 50.6 $\pm$ 7.0 & 82.9 $\pm$ 6.5 &  82.6 $\pm$ 3.2 &  78.7 $\pm$ 4.8 &67.5 $\pm$ 6.2 & \textbf{90.2 $\pm$ 1.4} \\
\midrule
Avg   & 58.5 $\pm$ 6.6 & 72.2 $\pm$ 9.6 & 82.6 $\pm$ 3.8  & 81.4 $\pm$ 1.8 &  71.8  $\pm$ 6.7  & \textbf{87.2 $\pm$ 2.7} \\
\midrule  \vspace{-0.2cm} \\
\multicolumn{7}{c} {FashionMNIST } \\
\midrule
T-shirt & 50.9 $\pm$ 5.5 &   86.8 $\pm$ 3.3 &  83.5 $\pm$ 2.1 & \textbf{97.5 $\pm$ 0.5}  &  79.7 $\pm$ 3.0 & 86.5 $\pm$ 1.1 \\
Trouser & 52.9 $\pm$ 12.7 & 94.4 $\pm$ 4.0  & 63.6 $\pm$ 4.6   & 80.2 $\pm$ 0.75 & 97.5 $\pm$ 1.7 & \textbf{99.0 $\pm$ 0.2}    \\
Pullover & 69.2 $\pm$ 5.8  & 81.2 $\pm$ 3.4 &  66.7 $\pm$ 2.8 & \textbf{90.1 $\pm$ 1.6}  & 89.2 $\pm$ 1.0  & 86.5 $\pm$ 1.1 \\
Dress   &  36.9 $\pm$ 8.5 & 88.1 $\pm$ 3.6  &  63.1 $\pm$ 0.8 &  91 $\pm$ 1.7  &  87.3 $\pm$ 1.5  & \textbf{91.7 $\pm$ 1.3}  \\
Coat  &   67.9 $\pm$7.6 &  84.7 $\pm$ 3.5 &  63.6 $\pm$ 4.6  & 88.5 $\pm$ 4.3  & 86.9 $\pm$ 0.9  & \textbf{88.9 $\pm$ 1.2} \\
Sandals & 54.1 $\pm$ 8.6 & 83.0 $\pm$ 12.4  &  64.9 $\pm$ 6.4  & 86.3 $\pm$ 1.0 & 72.5 $\pm$ 13.1  & \textbf{89.1 $\pm$ 1.6} \\
Shirt &  55.6 $\pm$ 7.8 & 74.8 $\pm$ 3.8   & 75.1 $\pm$ 3.9   &  \textbf{79.5 $\pm$ 2.5}   &  76.3 $\pm$ 2.0  &78.5 $\pm$ 0.8 \\
Sneaker &  56.8 $\pm$ 7.8  & 93.3 $\pm$ 1.4 & 59.1 $\pm$ 3.9  &  97.8 $\pm$ 0.4  &  96.3 $\pm$ 1.2  & \textbf{99.0 $\pm$ 0.2} \\
Bag &  56.1 $\pm$ 8.1  &  73.8 $\pm$ 10.2  &    72.4 $\pm$ 4.6 &  \textbf{98.4 $\pm$ 0.3}  & 77.9 $\pm$2.5   &94.5 $\pm$ 0.4 \\
Ankle-Boot & 60.3 $\pm$ 12.8   & 85.3 $\pm$ 3.7  &   71.2 $\pm$ 1.1 & 89.6 $\pm$ 0.7  &  97.5 $\pm$ 1.0 & \textbf{98.0 $\pm$ 0.6} \\
\midrule
Avg   & 56.1 $\pm$ 8.5   &  84.5 $\pm$ 4.9  &   68.3 $\pm$ 3.5 &   89.9 $\pm$ 1.4  & 86.1 $\pm$ 2.8  & \textbf{91.2 $\pm$ 0.9}  \\
\bottomrule \\
\end{tabular}}
\caption{Average AUC  (with standard deviation) for \textbf{Ten-Shot} anomaly detection experiments on Paris, CIFAR10, FashionMNIST and MNIST datasets. 
}
\label{tab:ten_shot}

\end{table*}

 \begin{table*}
 \centering
 \begin{tabular}{ccccccc}
 \toprule
 Class  & PatchSVDD & DROCC  & DeepSVDD & GEOM & GOAD & \textbf{Ours} \\
 \toprule
 \multicolumn{7}{c} {CIFAR10 (50-Shot)} \\
 \midrule
 Plane &3$6.7\pm6.7$  & $\bf{76.2 \pm 2.6} $ & $57.3 \pm 2.6$ & $67.8 \pm 2.9$ & $55.6 \pm 6.4$  & 75.9 $\pm$ 5.9 \\
 Car   & $65.5\pm3.6$  &  $44.7 \pm 3.0$ & $64.1 \pm 1.6$ & $82.4 \pm 1.3$ & $54.3 \pm 7.9$  & \textbf{86.2 $\pm$ 1.1}  \\
 Bird  & $38.1\pm2.1 $& $\bf{66.3 \pm 1.2} $ & $46.5 \pm 2.2$ & $60.3 \pm 3.1$ & $52.0 \pm 2.1$  &  57.3 $\pm$ 1.9 \\
 Cat   & $51.3 \pm 3.9$ & $\bf{61.4 \pm 4.0}$ & $58.5 \pm 2.2$ & $59.6 \pm 5.1$ & $49.8 \pm 0.6$ & 60.5 $\pm$ 1.0 \\
 Deer & $46.3 \pm 4.2$ & 5$8.6 \pm 2.9$ & $53.7 \pm 3.1$ & $57.4 \pm 5.2$ & $50.4 \pm 0.9$  & \textbf{64.5 $\pm$ 1.0} \\
 Dog   & $49.4 \pm 3.4$ & $63.3 \pm 5.4$ & $61.7 \pm 2.3$  & $68.6 \pm 2.6$  & $51.8 \pm 3.8$ & \textbf{74.7 $\pm$ 2.1} \\
 Frog  & $54.0 \pm 5.6$  & $45.8 \pm 2.6$ & $58.0 \pm 2.7$ & $64.8 \pm 2.8$ & $50.7 \pm 1.0$ & \textbf{73.2 $\pm$ 1.6} \\
 Horse & $55.4 \pm 3.1$ & $47.4 \pm 2.6$ & $62.3 \pm 3.2$ & $72.4 \pm 3.1$ & $52.7 \pm 5.4$ &  \textbf{74.5 $\pm$ 3.3} \\
 Ship  & $44.0 \pm 2.4$ & $74.7 \pm 2.7$ & $75.1 \pm 1.1$ & $81.4 \pm 1.7$ & $59.3 \pm 12.1$ &  \textbf{85.6 $\pm$ 0.6}\\
 Truck & $60.7 \pm 4.7$ & $37.4 \pm 5.12$ & $71.9 \pm 1.9$ & $\bf{81.1 \pm 2.1}$ &  $60.4 \pm 11.0$ &  76.8 $\pm$ 1.2\\
 \midrule
 Avg &  $50.1 \pm 4.0 $ & $57.6 \pm 3.2 $ & $60.9 \pm 2.3 $  &  $69.6 \pm 3.0 $ & $53.7 \pm 5.1$ &  \textbf{72.9 $\pm$  2.0}  \\
 \midrule \vspace{-0.2cm} \\
  \multicolumn{7}{c} {CIFAR10 (80-Shot)} \\
 \midrule
 Plane &  34.0 $\pm$ 4.5  & \textbf{79.0 $\pm$ 0.6} & 60.9 $\pm$ 2.1  & 69.9 $\pm$ 1.6 & 52.1 $\pm$ 4.3 &74.8 $\pm$ 0.3 \\
 Car   & 63.8 $\pm$ 6.9 & 43.2 $\pm$ 2.1 & 60.1 $\pm$ 0.8 & 85.3 $\pm$ 0.8 & 59.2 $\pm$ 11.3 & \textbf{88.0 $\pm$ 1.5} \\
 Bird & 40.0  $\pm$ 1.6 & \textbf{68.2 $\pm$ 0.3} & 44.6 $\pm$ 1.2  & 60.8 $\pm$ 2.4 & 50.7 $\pm$ 1.4 & 62.4 $\pm$ 1.2\\
 Cat   & 54.9 $\pm$ 1.3 & 55.7 $\pm$ 4.0 & 58.7 $\pm$ 0.2 & \textbf{62.9 $\pm$ 1.3} & 53.8 $\pm$ 4.6 &60.1 $\pm$ 1.4 \\
 Deer &  50.0$\pm$ 1.8 &57.2 $\pm$ 3.4 & 56.3 $\pm$ 0.8 & 62.7 $\pm$ 0.3 & 50.1 $\pm$ 2.4 & \textbf{66.1 $\pm$ 0.5} \\
 Dog   &  48.2 $\pm$ 3.2 & 64.4 $\pm$ 1.9 & 60.9 $\pm$ 1.7   & 76.5 $\pm$ 1.2 & 52.5 $\pm$ 5.0 & \textbf{78.4 $\pm$ 1.1}\\
 Frog  &  57.0 $\pm$ 2.3  & 50.9 $\pm$ 6.9 & 58.5 $\pm$ 2.5 & 69.9 $\pm$ 4.0 & 51.5 $\pm$ 7.1 & \textbf{75.3 $\pm$ 5.4} \\
 Horse & 56.7 $\pm$ 1.8 & 47.6 $\pm$ 2.1& 60.9 $\pm$ 0.1 & 79.9 $\pm$ 0.4 & 52.1 $\pm$ 3.9 & \textbf{82.3 $\pm$ 0.2}\\
 Ship  &  44.0 $\pm$ 3.5  & 77.0 $\pm$ 2.1 & 74.8 $\pm$ 0.1 & 84.0 $\pm$ 1.2 & 70.4 $\pm$ 10.5 & \textbf{87.4 $\pm$ 0.8} \\
 Truck & 61.2  $\pm$ 2.9  & 42.4 $\pm$ 1.1  & 72.1 $\pm$ 1.7 & \textbf{83.4 $\pm$ 0.3}  & 69.7 $\pm$ 9.9  &81.2 $\pm$ 0.6\\
 \midrule
 Avg  & 51.0 $\pm$ 3.0 & 58.5 $\pm$ 2.5 & 60.8 $\pm$ 1.1 & 73.5 $\pm$ 1.4 & 56.2 $\pm$ 6.1 & \textbf{75.6 $\pm$ 1.3}  \\
 \bottomrule \\
 
 \end{tabular}

 \caption{Average AUC  (with standard deviation) for \textbf{50-shot} and \textbf{80-shot} anomaly detection experiments on CIFAR10.
 }
 \label{tab:50_shot_anomaly}
 \end{table*}

\begin{table*}
\centering
\scalebox{0.9}{%
\begin{tabular}{ccccccccc}
\toprule 
Class  & DifferNet & DROCC & PatchSVDD & DeepSVDD & GEOM & GOAD & \textbf{Ours1} & \textbf{Ours2} \\ 
 \toprule
 \multicolumn{9}{c} {MVTec (One-Shot)} \\
\midrule
Bottle & \textbf{98.2 $\pm$ 0.4} & 67.2 $\pm$ 6.6 & 60.9 $\pm$ 12.3 & 16.6 $\pm$ 5.3& 79.0 $\pm$ 3.5 & 51.6 $\pm$ 14.0 & 76.3 $\pm$ 6.9 & 85.0 $\pm$ 3.7 \\
Cable & \textbf{76.6 $\pm$ 5.9} & 68.1 $\pm$ 4.3& 58.8 $\pm$ 4.5 & 39.0 $\pm$ 3.5 & 64.2 $\pm$ 1.3 & 47.9 $\pm$ 2.4& 72.3 $\pm$ 3.7 &  61.1 $\pm$ 7.8 \\
Capsule &57.7 $\pm$ 4.6& 50.2 $\pm$ 6.4& 57.9 $\pm$ 12.1& 44.8 $\pm$ 4.4& 55.4 $\pm$ 2.6 &51.2 $\pm$ 3.7 &56.0 $\pm$ 8.4 & \textbf{62.6 $\pm$ 6.7} \\
Carpet &61.5 $\pm$3.0 &71.9 $\pm$ 10.6&45.5 $\pm$ 18.8 & 41.2$\pm$ 18.2 & 55.0 $\pm$ 10.1 &48.1 $\pm$ 1.9 & \textbf{72.7 $\pm$ 6.7} &  \textbf{83.7 $\pm$ 8.7} \\
Grid &59.2 $\pm$ 5.1 & 50.0 $\pm$ 4.6&37.2 $\pm$ 12.2 & 79.7 $\pm$ 8.6& 40.1 $\pm$ 13.1 & 9.4 $\pm$ 6.8&73.2 $\pm$ 9.8 & \textbf{87.1 $\pm$ 5.0} \\
Hazelnut & \textbf{90.7 $\pm$ 2.7} & 66.4 $\pm$ 7.6 & 46.7 $\pm$ 16.1& 29.1 $\pm$ 4.3& 47.8 $\pm$ 3.6 & 47.6 $\pm$ 3.2 & 82.4 $\pm$ 8.7 & 66.5 $\pm$ 9.2 \\
Leather & 83.4 $\pm$ 1.9& 79.1 $\pm$ 6.5& 61.9 $\pm$ 15.6& 48.0 $\pm$ 3.2&  33.2 $\pm$ 0.5 & 58.1 $\pm$ 6.8& \textbf{98.2 $\pm$ 0.9} & 97.6 $\pm$ 1.1 \\
Metalnut & 44.4 $\pm$ 8.0& 51.9 $\pm$ 3.6 & 50.4 $\pm$ 13.1 &42.6 $\pm$ 14.7 & 52.3 $\pm$ 4.2 &7.2 $\pm$ 6.5 & \textbf{66.0 $\pm$ 11.0}  & 60.3 $\pm$ 8.6 \\
Pill & 71.7 $\pm$ 4.4  & \textbf{72.5 $\pm$ 4.0} & 57.6 $\pm$ 8.1& 33.5 $\pm$ 4.0& 67.0 $\pm$ 2.3 & 62.5 $\pm$ 8.1& 56.5 $\pm$ 9.6 &  66.5 $\pm$ 7.0 \\
Screw & 61.8 $\pm$ 7.7 & 57.7 $\pm$ 9.0 & 53.7 $\pm$ 18.2& 70.1 $\pm$ 10.8& 34.7 $\pm$ 11.1 & 6.3 $\pm$10.0& \textbf{93.5 $\pm$ 6.2} &  92.8 $\pm$ 6.0\\
Tile & \textbf{87.3 $\pm$ 2.6} & 65.6 $\pm$ 2.0 & 57.3 $\pm$ 4.7 &40.7 $\pm$ 2.8 & 61.0 $\pm$ 2.8 & 6.0 $\pm$ 5.4 & 80.2 $\pm$ 8.2 &  84.4 $\pm$ 3.8 \\
Toothbrush &52.1 $\pm$ 2.3 & \textbf{68.9 $\pm$ 4.5} & 63.7 $\pm$ 6.1 & 35.5 $\pm$ 1.5& 65.7 $\pm$ 6.5 & 54.4$\pm$5.4 &67.3 $\pm$ 4.7& 64.7 $\pm$ 11.1 \\
Transistor & 47.0 $\pm$ 6.5& 59.9 $\pm$ 3.3& \textbf{66.7 $\pm$ 14.5} &32.8 $\pm$ 4.3 & 58.1 $\pm$ 1.5 & 61.7 $\pm$ 4.4& 66.1 $\pm$ 7.7 & 62.7 $\pm$ 6.8 \\
Wood & \textbf{96.0 $\pm$ 2.2} & 70.6 $\pm$ 14.4 & 55.7 $\pm$ 18.4 & 44.0 $\pm$ 16.4& 52.3 $\pm$ 1.1 & 41.8 $\pm$ 6.5& 89.0 $\pm$ 4.2 & 85.5 $\pm$ 7.9 \\
Zipper &52.7 $\pm$ 3.7 &  49.6 $\pm$ 7.5 & 69 $\pm$ 5.4 & 34.9 $\pm$ 2.8& 58.3 $\pm$ 2.8   & 56.8 $\pm$ 4.0& \textbf{67.8 $\pm$ 6.4} & \textbf{73.2 $\pm$ 7.7} \\
\midrule
Avg &69.4 $\pm$ 4.1 & 63.3 $\pm$ 6.3 & 56.2 $\pm$ 12.0 & 42.1 $\pm$ 7.0  & 54.9 $\pm$ 4.5 & 44.0 $\pm$ 5.9 & \textbf{74.5 $\pm$ 6.9}& \textbf{75.6 $\pm$ 6.7} \\
\midrule \vspace{-0.2cm}\\

\multicolumn{9}{c} {MVTec (Five-Shot)} \\
\midrule
Bottle & \textbf{98.4 $\pm$ 0.2} & 68.1 $\pm$ 2.6 & 61.1 $\pm$ 12.4 & 15.7 $\pm$ 2.8 & 80.0 $\pm$ 1.2 & 51.7 $\pm$ 10.4 &  74.1 $\pm$ 7.8 & 90.8 $\pm$ 3.7 \\
Cable & \textbf{81.3 $\pm$ 2.0} & 68.7 $\pm$ 2.7 & 49 $\pm$ 3.9 & 32.8 $\pm$ 4.9 & 61.1 $\pm$ 3.1 & 46.3 $\pm$ 4.4 &75.2 $\pm$ 4.8 & 76.1 $\pm$ 4.0 \\
Capsule & 59.0 $\pm$ 2.2 & 53.2 $\pm$ 5.1 &  55.1 $\pm$ 3.4 &  45.3 $\pm$ 4.7 & 60.0 $\pm$ 2.3 &  47.7 $\pm$ 5.9 & 52.6 $\pm$ 6.5& \textbf{64.9 $\pm$ 5.6} \\
Carpet & 62.0 $\pm$ 2.2 & 71.6 $\pm$ 10.9 & 46.5 $\pm$ 4.1 &  47.7 $\pm$ 10.5 & 42.2 $\pm$ 6.7 &  44.2 $\pm$ 6.9 & \textbf{73.3 $\pm$ 7.6} & 65.2 $\pm$ 6.4 \\
Grid & 56.7 $\pm$ 3.9 & 37.3 $\pm$9.7 & 41.7 $\pm$ 22.1 & 76.0 $\pm$ 11.1 & 36.8 $\pm$ 7.2 & 21.3 $\pm$ 16.4  & \textbf{76.0 $\pm$ 4.9} & \textbf{82.4 $\pm$ 9.7} \\
Hazelnut & \textbf{93.8 $\pm$ 1.0} & 70.0 $\pm$ 10.9 & 58.6 $\pm$ 17.4 & 27.7 $\pm$ 4.6 & 31.7 $\pm$ 8.2 & 52.5 $\pm$ 3.5  & 76.8 $\pm$ 8.3 & 84.5 $\pm$ 8.8 \\
Leather & 83.7 $\pm$ 0.8&  70.4 $\pm$ 7.1 & 61.6 $\pm$ 15.4 & 43.0 $\pm$ 2.0 & 33.3 $\pm$ 0.2 & 53.2 $\pm$ 10.3  &  \textbf{99.0 $\pm$ 0.3} & \textbf{98.2 $\pm$ 0.9} \\
Metalnut &47.2 $\pm$ 3.2 & 59.7 $\pm$ 6.2 & 48.8 $\pm$ 9.1 &  52.9 $\pm$ 6.6 & 36.8 $\pm$ 4.3 & 59.4 $\pm$ 5.6 & \textbf{69.4 $\pm$ 11.4} & \textbf{76.4 $\pm$ 6.5} \\
Pill & \textbf{79.4 $\pm$ 4.4} &  74.4 $\pm$ 3.5 & 57.5 $\pm$ 10.6 & 34.4 $\pm$ 3.5 & 59.1 $\pm$ 3.1 & 61.5 $\pm$ 11.0 & 51.2 $\pm$ 6.8& 63.6 $\pm$ 4.1 \\
Screw &73.7 $\pm$ 5.1 & 58.3 $\pm$ 2.3&  43.4 $\pm$ 15.1 &  69.5 $\pm$ 3.8 & 18.5 $\pm$ 5.1 & 9.3 $\pm$ 13.6 & \textbf{97.7 $\pm$ 3.2} &  \textbf{74.8 $\pm$ 1.3}\\
Tile & \textbf{91.1 $\pm$ 1.4} &65.7 $\pm$ 3.1 & 49.5 $\pm$ 3.0 &  32.4 $\pm$ 3.2 & 56.9 $\pm$ 11.1 & 58.6 $\pm$ 3.9 & 89.0 $\pm$ 4.5  & 81.0 $\pm$ 4.4 \\
Toothbrush &57.3 $\pm$ 3.6 & 67.6 $\pm$ 3.6 & 68.3 $\pm$ 11.8 & 34.9 $\pm$ 6.7 & 72.2 $\pm$ 2.1 & 45.3 $\pm$ 4.5 & \textbf{72.7 $\pm$ 8.1} & 64.2 $\pm$ 7.3 \\
Transistor &55.7 $\pm$ 3.9 & 67.2 $\pm$ 4.1 & 55.3 $\pm$ 9.9 & 30.4 $\pm$ 2.6 & 59.4 $\pm$ 2.9 &  62.8 $\pm$4.0 & \textbf{78.2 $\pm$ 4.2} & \textbf{76.2 $\pm$ 3.9} \\
Wood & \textbf{96.4 $\pm$ 1.9} & 77.7 $\pm$ 11.9 & 69.4 $\pm$ 14.6 & 11.0 $\pm$ 7.3 & 66.0 $\pm$ 9.8 & 37.4 $\pm$ 9.8 & 84.5 $\pm$ 3.6 &  96.2 $\pm$ 1.8 \\
Zipper & 46.1 $\pm$ 3.7 & 45.2 $\pm$ 6.1 & 63.9$\pm$6.5 & 34.4 $\pm$ 4.6 & 59.2 $\pm$ 6.2 &  54.1 $\pm$ 8. & 61.8 $\pm$ 7.2 &  \textbf{73.3 $\pm$ 10.7} \\
\midrule
Avg & 72.1 $\pm$ 2.6 & 63.7 $\pm$ 6.0 & 55.3 $\pm$ 10.6  & 39.2 $\pm$ 5.3 & 51.5 $\pm$ 4.9 &  47.0 $\pm$ 7.9 & \textbf{75.4 $\pm$ 5.9} & \textbf{77.9 $\pm$ 5.3} \\
\midrule \vspace{-0.2cm} \\

\multicolumn{9}{c} {MVTec (Ten-Shot)} \\
\midrule
Bottle & \textbf{98.2 $\pm$ 0.4}  & 67.7 $\pm$ 5.1 & 65.3 $\pm$ 9.6 & 17.6 $\pm$ 3.0 & 80.1 $\pm$ 2.5  & 86.9 $\pm$ 4.5 & 81.9 $\pm$ 6.1  & 90.5 $\pm$ 3.1 \\
Cable & \textbf{82.3 $\pm$ 1.5} & 69.4 $\pm$ 3.0 &  51.1 $\pm$ 7.7 & 32.6 $\pm$ 2.5 & 64.4 $\pm$ 0.8 & 46.0 $\pm$ 9.9 & 73.9 $\pm$ 4.2  & 77.6 $\pm$ 3.9 \\
Capsule & 58.0 $\pm$ 2.1 & 51.8 $\pm$ 6.3 & 64.4 $\pm$ 11.1 & 44.7 $\pm$ 2.9 & \textbf{65.9 $\pm$ 0.8} & 47.3 $\pm$ 2.0 & 55.8 $\pm$ 7.7 & 59.3 $\pm$ 8.4 \\
Carpet & 61.8 $\pm$ 1.5 & \textbf{75.1 $\pm$ 16.4} & 49.4 $\pm$ 7.4 &  40.0 $\pm$ 11.6 & 41.4 $\pm$ 7.0 & 50.9 $\pm$ 8.5 & 66.9 $\pm$ 9.6  & 63.9 $\pm$ 6.8 \\
Grid &58.5 $\pm$ 2.1 &  37.5 $\pm$ 17.1 & 49.8 $\pm$ 11.1 & 67.1 $\pm$ 10.6 & 10.3 $\pm$ 6.7 &  54.0 $\pm$ 7.1 & \textbf{71.0 $\pm$ 8.6} &  \textbf{79.0 $\pm$ 5.9} \\
Hazelnut & \textbf{93.2 $\pm$ 1.3} & 72.7 $\pm$ 11.9  & 37.9 $\pm$ 12.0 & 30.5 $\pm$ 5.2 & 45.1 $\pm$ 1.6 & 49.6 $\pm$ 2.7 & 72.1 $\pm$ 8.2  & 79.3 $\pm$ 11.3 \\
Leather & 83.4 $\pm$ 0.9 & 79.1 $\pm$ 13.8 & 49.3 $\pm$ 15.9 & 43.5 $\pm$ 2.8 & 32.7 $\pm$ 0.8 &  61.2 $\pm$ 5.2 & \textbf{99.1 $\pm$ 0.2} & 98.5 $\pm$ 0.5 \\
Metalnut & 53.4 $\pm$ 7.4 & 59.1 $\pm$ 6.6 & 62.3 $\pm$ 12.5 & 52.4 $\pm$ 3.9 & 49.3 $\pm$ 1.4 & 58.6 $\pm$ 6.7 & 60.4 $\pm$ 11.8  &  74.0 $\pm$ 8.4 \\
Pill & \textbf{81.8 $\pm$ 3.5} & 77.6 $\pm$ 3.6 & 65.2 $\pm$ 8.4 &  39.1 $\pm$ 3.9 & 56.1 $\pm$ 1.2 & 64.1 $\pm$ 3.0 & 57.4 $\pm$ 10.4 & 66.5 $\pm$ 7.0 \\
Screw & 78.3 $\pm$ 4.3 & 84.2 $\pm$ 19.8 &  28.8 $\pm$ 21.3 & 65.2 $\pm$ 4.3 & 8.5 $\pm$  6.3 & 66.7 $\pm$ 0.8 & \textbf{93.9 $\pm$ 8.4} & 75.7 $\pm$ 19.0 \\
Tile & \textbf{91.3 $\pm$ 1.2} & 64.8 $\pm$ 4.2 & 49.0 $\pm$ 3.1 & 26.0 $\pm$ 5.0 & 62.0 $\pm$ 0.3 & 54.3 $\pm$ 3.5 & 87.6 $\pm$ 5.5   & 81.4 $\pm$ 6.9 \\
Toothbrush & 57.5 $\pm$ 4.0 & 67.9 $\pm$ 3.3 &  67.3 $\pm$ 9.6 & 38.2 $\pm$ 7.6  & 71.5 $\pm$ 0.4 & 51.3 $\pm$ 8.6 & \textbf{78.9 $\pm$ 8.5} &  69.5 $\pm$ 7.7 \\
Transistor &54.6 $\pm$ 3.7 & 72.5 $\pm$ 3.6 & 60.3 $\pm$ 6.2 & 24.6 $\pm$ 4.5 &58.9 $\pm$ 3.1  & 56.0 $\pm$ 8.4 & \textbf{74.9 $\pm$ 3.7}  & 79.2 $\pm$ 4.7 \\
Wood & \textbf{96.2 $\pm$ 1.9} &84.0 $\pm$ 8.2  & 47.9 $\pm$ 12.3 &  18.3 $\pm$ 11.6 & 67.7 $\pm$ 5.5 & 37.4 $\pm$ 5.9 & 85.0 $\pm$ 5.9 & 95.8 $\pm$ 1.1 \\
Zipper & 55.2 $\pm$ 6.1 & 50.0 $\pm$ 6.7 &  66.7 $\pm$ 4.8 & 36.1 $\pm$ 4.5 & 60.9 $\pm$ 2.2 &  53.1 $\pm$ 12.3 & \textbf{72.8 $\pm$ 6.5} &  \textbf{80.4 $\pm$ 5.9} \\
\midrule
Avg & 73.6 $\pm$ 2.8 & 67.6 $\pm$ 8.6 & 54.3 $\pm$ 10.2 & 38.4 $\pm$ 5.6 & 51.6 $\pm$ 2.7 & 55.8 $\pm$ 5.9 & \textbf{75.4 $\pm$ 7.0} & \textbf{78.0 $\pm$ 6.7} \\
\bottomrule \\

\end{tabular}}

\caption{ Average AUC  (with standard deviation) for \textbf{One-Shot}, \textbf{Five-Shot}  and \textbf{Ten-Shot} defect detection experiments on MVTec dataset. \textbf{Ours1} refers to our method where the standard set of transformations are used, as for anomaly detection. For a fair comparison with DifferNet, we also consider \textbf{Ours2}, where only the four rotation are used, as in DifferNet. In the one-shot case, we report the results of using 5\% of the patches, while in five-shot and ten-shot case we report the results of using 10\% of the patches. The full results of using different percentage of patches are given in Tab.~\ref{tab:defect_detection_ablation_1shot}}.
\label{tab:one_shot_defect}
\end{table*}

\begin{table*}
\centering
 \renewcommand\theadfont{}
\begin{tabular}{ccccccccc}
\toprule
Class& \textbf{Ours} & (a)  & (b) & (c)  &  (d)  & (e) &  (f) & (g)\\
\toprule
\multicolumn{8}{c} {CIFAR10 (One-Shot Ablation) } \\
\midrule
Plane & \textbf{67.2 $\pm$ 5.8}  &  58.9 $\pm$ 12.5 & 65.2 $\pm$ 10.6 &  65.2 $\pm$ 5.6 & 59.9 $\pm$ 9.9  &   60.1 $\pm$ 5.9 &  27.0 $\pm$ 0.4 & 38.2 $\pm$ 3.9 \\
Car  &   \textbf{65.6 $\pm$ 5.9}  &  61.6 $\pm$ 7.8   & 65.5 $\pm$ 3.5 &  58.3 $\pm$ 3.6 & 55.0 $\pm$ 8.6 & 63.6 $\pm$ 5.8  & 59.1 $\pm$ 1.4 & 57.6 $\pm$ 4.2\\
Bird & 55.9 $\pm$ 5.7 &  52.6 $\pm$ 6.3 & \textbf{56.0 $\pm$ 4.2}  &  54.2 $\pm$ 3.2 & 52.9 $\pm$ 5.9 &  48.9 $\pm$ 6.8  & 44.7 $\pm$ 1.3& 46.3 $\pm$ 2.1\\
Cat &  58.9 $\pm$ 6.2 &  53.8 $\pm$ 8.0 & 55.7 $\pm$ 3.2  &  56.8 $\pm$ 3.6 & 48.2 $\pm$ 6.6 & 54.3 $\pm$ 5.4 & 54.9 $\pm$ 1.0& \textbf{66.4 $\pm$ 3.1} \\
Deer & 67.2 $\pm$ 4.5 &   61.9 $\pm$ 6.8 & 55.7 $\pm$ 8.9 &  56.5 $\pm$  10.1 & \textbf{67.8 $\pm$ 2.6} & 53.6 $\pm$ 8.1 & 51.4 $\pm$ 2.8 & 67.3 $\pm$ 5.5\\
Dog  &63.7 $\pm$ 7.7 &  61.0 $\pm$ 7.8 & 53.0 $\pm$ 4.1 &   60.0 $\pm$ 3.4 & 55.8 $\pm$ 7.8 & 57.5 $\pm$ 7.6 &  50.0 $\pm$ 2.8 & \textbf{65.9 $\pm$ 5.1} \\
Frog &  \textbf{70.2 $\pm$ 5.1} &  65.1 $\pm$ 9.9 & 56.4 $\pm$ 8.1  &   62.5 $\pm$ 4.2 & 62.3 $\pm$ 9.6& 57.5 $\pm$ 8.1 &  58.0 $\pm$ 2.1 & 68.2 $\pm$ 4.2\\
Horse & \textbf{63.8 $\pm$ 5.2} &  61.8 $\pm$ 7.8 & 53.7 $\pm$ 4.1  &  59.4 $\pm$ 3.5 & 54.6 $\pm$ 7.6& 59.7 $\pm$ 7.8 & 51.8 $\pm$ 0.5 & 39.8 $\pm$ 3.4\\
Ship & \textbf{71.3 $\pm$ 7.2} &  70.4 $\pm$ 9.5 & 65.1 $\pm$ 10.2 &  62.6 $\pm$ 7.5 & 69.5 $\pm$ 9.4 &  58.1 $\pm$ 8.5 & 33.9 $\pm$ 2.3 & 65.1 $\pm$ 3.5\\
Truck & 65.3 $\pm$ 5.2 & 60.3 $\pm$ 8.8 & 64.8 $\pm$ 4.1  &  61.2 $\pm$ 7.2  & 50.0 $\pm$ 6.2 5&  59.1 $\pm$ 4.8 & 46.5 $\pm$ 3.3& \textbf{74.1 $\pm$ 3.7} \\
\midrule
Avg  & \textbf{64.9 $\pm$ 5.9} &  60.7 $\pm$ 8.5 &  59.1 $\pm$ 6.1 &   59.7 $\pm$ 5.2 & 57.6 $\pm$ 7.4 &  57.3 $\pm$ 6.9 & 47.7 $\pm$ 1.8 & 58.8  $\pm$ 3.9 \\

\midrule \vspace{-0.2cm} \\
\multicolumn{8}{c} {CIFAR10 (Five-Shot Ablation)} \\
\midrule
Plane &  \textbf{69.2 $\pm$ 2.8} &  68.1 $\pm$ 2.7 & 65.1 $\pm$ 8.4  &  61.4 $\pm$ 3.9 &   57.8 $\pm$ 7.8 &  65.9 $\pm$ 3.9  & 25.9 $\pm$ 0.3 & 50.7 $\pm$ 3.1\\
Car&  \textbf{77.0 $\pm$ 1.8}   & 75.2 $\pm$ 4.6 & 59.2 $\pm$ 8.6  & 70.2 $\pm$ 2.7  & 56.7 $\pm$ 4.9  &  70.6 $\pm$ 5.1  &  60.0 $\pm$ 1.0 & 73.1 $\pm$ 2.7\\
Bird & 58.4 $\pm$ 2.3 &  52.7 $\pm$ 2.2  & 58.4 $\pm$ 3.3 & 56.2 $\pm$ 2.4   & \textbf{58.9 $\pm$ 6.5}  & 51.5 $\pm$ 5.1 &  45.2 $\pm$ 0.6 & 50.4 $\pm$ 2.0\\
Cat  &  \textbf{58.7 $\pm$ 4.3} & 55.1 $\pm$ 4.9 & 53.7 $\pm$ 3.2  &  58.2 $\pm$ 4.2  & 50.4 $\pm$ 7.6  & 53.8 $\pm$ 5.9 &  55.7 $\pm$ 1.0 & 56.3 $\pm$ 2.5\\
Deer & \textbf{66.4 $\pm$ 4.3} & 63.1 $\pm$ 4.2  & 66.2 $\pm$ 5.5 &  61.3 $\pm$ 4.9   & 64.9 $\pm$ 4.9  &  60.2 $\pm$ 3.5  &  50.9 $\pm$ 0.7 & 59.4 $\pm$ 5.0\\
Dog  & 61.8 $\pm$ 3.2&  57.4 $\pm$ 9.6 & 53.5 $\pm$ 2.9   &  61.2 $\pm$ 3.9  & 50.5 $\pm$ 8.3 &  \textbf{64.1 $\pm$ 3.3}   & 51.4 $\pm$ 1.5 & 60.9 $\pm$ 4.0\\
Frog &  \textbf{72.6 $\pm$ 4.4} & 66.1 $\pm$ 4.7   & 67.1 $\pm$ 8.3  & 66.3 $\pm$ 6.8 & 65.4 $\pm$ 0.3 &  64.1 $\pm$ 2.5  & 57.7 $\pm$ 0.8 & 69.1 $\pm$ 4.1\\
Horse &   \textbf{68.6 $\pm$ 2.8} & 67.6 $\pm$ 5.9  & 55.3 $\pm$ 3.0  & 63.3 $\pm$ 2.6  & 55.5 $\pm$ 11.4  &  66.9 $\pm$ 5.7 & 51.8 $\pm$ 0.4 & 66.9 $\pm$ 3.0\\
Ship  & \textbf{80.2 $\pm$ 3.2} & 76.2 $\pm$ 5.2 & 66.2 $\pm$ 6.2 & 67.5 $\pm$ 6.0 & 65.3 $\pm$ 6.8  &72.2 $\pm$ 5.5 &  34.1 $\pm$ 1.2 & 76.4 $\pm$ 3.2\\
Truck & 62.1 $\pm$ 3.4 & 67.8 $\pm$ 3.8  & 55.3 $\pm$ 7.2  & 66.5 $\pm$ 3.7 & 53.0 $\pm$ 3.2 &68.7 $\pm$ 5.9 & 47.4 $\pm$ 1.5 & \textbf{74.3 $\pm$ 3.1} \\
\midrule
Avg  & \textbf{67.5 $\pm$ 3.4} & 64.9 $\pm$ 4.8 &  60.0 $\pm$ 5.7  & 63.4  $\pm$ 4.1  &  57.8 $\pm$ 6.2   & 63.8 $\pm$ 4.6  & 48.0 $\pm$ 0.9 & 63.7  $\pm$ 3.3\\
\bottomrule \\

\end{tabular}
\caption{ \textbf{Ablation analysis} for \textbf{One-Shot} and \textbf{Five-Shot} anomaly detection, as described in the main text, Sec.~4.3, Tab.~1. Our method relies on three components: (1) a generative model, (2) its hierarchical multi-scale nature, and (3) a transformation-discriminating component. We assess the contribution of these components separately. The columns of the table represent different variants: (a) no generative component, (b) transformations not applied discriminatively, (c) as for (b), but where augmentations are applied before passing real and generated images to the discriminator. (d) a single scale of the hierarchy where small patches are considered (image size set to $100 \times 100$), (e) a single scale of the hierarchy where large patches are considered (image size set to $20 \times 20$), (f) no component is used and the anomaly score is the MSE between the test image and the training image (average for each training image for five-shot). Finally, the last variant (g) trains a GEOM model on $6,000$ images sampled from our generative model that is trained on a one/five sample. }
\label{tab:ablation}
\end{table*}

 \begin{table*}
 \centering
 \scalebox{0.9}{
 
 \begin{tabular}{ccccccc}
 \toprule
 Fraction (\%)  & 1 & 5  & 10 & 20 & 50 & 100 \\
 \toprule
 \multicolumn{7}{c} {MVTec  (One-Shot)} \\
 \midrule
Bottle &  75.4 $\pm$ 12.6 & 85.0 $\pm$ 3.7 & 76.5 $\pm$ 9.0&  82.5 $\pm$ 9.0 & 81.6 $\pm$ 6.3 & 67.0 $\pm$  9.4 \\
Cable  & 57.4 $\pm$ 9.3   &61.1 $\pm$ 7.8 & 67.8 $\pm$ 3.6 & 59.7 $\pm$ 11.9 &62.0 $\pm$ 10.7 & 54.0 $\pm$ 10.6 \\
Capsule &  59.2 $\pm$ 11.4 &62.6 $\pm$ 6.7  &  59.7 $\pm$ 6.2 &61.9 $\pm$ 6.4  &57.5 $\pm$ 6.0 & 58.4 $\pm$ 7.9\\
Carpet   & 81.4 $\pm$ 7.7 & 83.7 $\pm$ 8.7 &  81.6 $\pm$ 9.2 & 84.4 $\pm$ 4.9 &80.2 $\pm$ 10.4 & 69.8 $\pm$ 8.2 \\
Grid &  91.3 $\pm$ 4.8 & 87.1 $\pm$ 5.0&   83.3 $\pm$ 7.1 &  82.6 $\pm$ 5.2 &71.7 $\pm$ 7.9& 58.7 $\pm$ 8.4\\
Hazelnut  &  67.0 $\pm$ 10.1 & 66.5 $\pm$ 9.2 & 69.3 $\pm$ 10.0 & 67.4 $\pm$ 8.4  & 61.6 $\pm$ 13.9 & 65.2 $\pm$ 10.1 \\
Leather  &  98.0 $\pm$ 1.1 & 97.6 $\pm$ 1.1 &  96.7 $\pm$ 1.8 & 95.4 $\pm$ 2.8 &93.7 $\pm$ 4.2 & 81.7 $\pm$ 11.6\\
Metal-nut &  69.4 $\pm$ 14.0  & 60.3 $\pm$ 8.6 &65.8 $\pm$ 9.9 & 64.9 $\pm$ 10.4 &61.9 $\pm$13.6 & 67.0 $\pm$ 9.8\\
Pill &  66.8 $\pm$ 5.9  & 66.5 $\pm$ 7.0 &  66.1 $\pm$ 6.9 & 64.3 $\pm$ 6.3 &64.7 $\pm$ 8.2 & 59.0 $\pm$ 7.4\\
Screw &  92.9 $\pm$ 6.4 & 92.8 $\pm$ 6.0 &  89.1 $\pm$ 6.9 & 89.9 $\pm$ 7.0 &87.7 $\pm$ 6.9 & 61.8 $\pm$ 6.9\\
Tile &  85.1 $\pm$ 3.0  & 84.4 $\pm$ 3.8 & 83.0 $\pm$ 8.9 & 84.2 $\pm$ 4.1 &79.1 $\pm$ 5.4 & 57.7 $\pm$ 4.4\\
Toothbrush &  61.9 $\pm$ 11.5 & 64.7 $\pm$ 11.1 &  57.5 $\pm$ 5.9& 58.4 $\pm$ 6.6 & 59.1 $\pm$ 6.4 & 56.9 $\pm$ 7.4\\
Transistor &  60.3 $\pm$ 7.3  & 62.7 $\pm$ 6.8 &  67.8 $\pm$ 5.8&63.9 $\pm$ 8.4 & 64.3 $\pm$ 8.4 & 66.8 $\pm$ 10.2\\
Wood &  82.0 $\pm$ 11.7 & 85.5 $\pm$ 7.9 & 81.7 $\pm$ 9.9& 82.9 $\pm$ 9.9 & 81.2 $\pm$ 11.4 & 71.7 $\pm$ 11.1\\
Zipper &  78.3 $\pm$ 8.7  & 73.2 $\pm$ 7.7 &   71.4 $\pm$ 9.7 & 72.5 $\pm$ 6.3 &72.7 $\pm$ 4.9 &63.6 $\pm$ 14.9  \\
 \midrule
 Avg &  75.1 $\pm$ 8.4 & 75.6 $\pm$ 6.7 & 74.5 $\pm$ 7.4 &  74.3 $\pm$ 7.2 &71.9 $\pm$ 8.3 & 63.9 $\pm$ 9.2 \\
 \midrule \vspace{-0.2cm} \\

 \multicolumn{7}{c} {MVTec  (Five-Shot)} \\
 \midrule
Bottle & 87.1 $\pm$ 6.5 &  90.2 $\pm$ 6.7 &  90.8 $\pm$ 3.7 &   88.3 $\pm$ 5.9  & 86.3 $\pm$ 9.6  & 84.4 $\pm$ 5.0  \\
Cable  &  71.6 $\pm$ 3.4 & 74.0 $\pm$ 3.4  & 76.1 $\pm$ 4.0   &  74.5 $\pm$ 4.1 &  74.7 $\pm$ 4.5  & 74.3 $\pm$ 4.5 \\
Capsule & 56.0 $\pm$ 6.1  & 60.2 $\pm$ 8.5   &  64.9 $\pm$ 5.6 & 57.0 $\pm$ 7.7 & 50.2 $\pm$ 6.8  & 51.1 $\pm$ 5.5 \\
Carpet  & 76.3 $\pm$ 9.1 &  72.9 $\pm$ 8.0  & 65.2 $\pm$ 6.4 &  59.7 $\pm$ 11.4 & 62.6 $\pm$ 10.9 & 46.6 $\pm$ 6.8   \\
Grid    &  90.3 $\pm$ 4.5  & 86.8 $\pm$ 4.7  &  82.4 $\pm$ 9.7  &  78.1 $\pm$ 7.9 & 68.2 $\pm$ 3.9 & 51.8 $\pm$ 6.2 \\
Hazelnut &  83.6 $\pm$ 4.2 &  82.2 $\pm$ 8.0 & 84.5 $\pm$ 8.8 &  76.7 $\pm$ 8.8 &  78.6 $\pm$ 7.7  & 70.4 $\pm$ 10.8 \\
Leather  &  98.8 $\pm$ 0.9  &  98.6 $\pm$ 0.7 &98.2 $\pm$ 0.9 & 96.9 $\pm$ 1.3  &  95.4 $\pm$ 2.2  &  76.6 $\pm$ 8.4\\
Metal-nut & 70.1 $\pm$ 8.7 &  72.1 $\pm$ 7.7 &  76.4 $\pm$ 6.5 &  70.0 $\pm$ 7.2 &  75.3 $\pm$ 8.0 & 80.5 $\pm$ 5.3 \\
Pill  &  66.4 $\pm$ 6.3  &  64.3 $\pm$ 7.5  &  63.6 $\pm$ 4.1 &  63.1 $\pm$ 8.6 &  60.6 $\pm$ 6.4  & 60.0 $\pm$ 4.3 \\
Screw  &  77.4 $\pm$ 8.3 & 76.4 $\pm$ 6.7  &  74.8 $\pm$ 1.3 & 64.1 $\pm$ 11.5  & 56.5 $\pm$ 13.1  &  43.1 $\pm$ 5.9\\
Tile  &   81.9 $\pm$ 6.3 &  80.4 $\pm$ 6.0 & 81.0 $\pm$ 4.4 & 75.4 $\pm$ 9.7  &  73.6 $\pm$ 9.4  &  50.2 $\pm$ 4.9\\
Toothbrush & 61.2 $\pm$ 6.2  & 62.2 $\pm$ 8.9  & 64.2 $\pm$ 7.3 & 60.9 $\pm$ 10.5 & 60.9 $\pm$ 7.7 &  62.2 $\pm$ 4.6 \\
Transistor  & 74.8 $\pm$ 4.5 &74.4 $\pm$ 6.4  &  76.2 $\pm$ 3.9 &  76.4 $\pm$ 6.6 &  80.2 $\pm$ 8.0 & 78.7 $\pm$ 5.1 \\
Wood    &  95.7 $\pm$ 1.9  &   96.4 $\pm$ 2.1 &96.2 $\pm$ 1.8  & 94.7 $\pm$ 3.0 & 93.0 $\pm$ 4.9  &  93.5 $\pm$ 6.5 \\
Zipper&  79.2 $\pm$ 8.1  & 74.8 $\pm$ 8.8 &  73.3 $\pm$ 10.7  &  75.0 $\pm$ 7.9 &  74.8 $\pm$ 6.6  &  78.6 $\pm$ 6.7\\

 \midrule
 Avg &   78.0 $\pm$ 5.7  &  77.7 $\pm$ 6.3  &   77.9 $\pm$ 5.3 &  74.0 $\pm$ 7.5 &   72.7 $\pm$ 7.3 & 66.8 $\pm$ 6.0 \\
 \midrule \vspace{-0.2cm} \\
 \multicolumn{7}{c} {MVTec (Ten-Shot) } \\
 \midrule
Bottle &  92.4 $\pm$ 3.3  & 92.7 $\pm$ 2.6  &  90.5 $\pm$ 3.1  & 90.7 $\pm$ 4.7 & 90.2 $\pm$ 2.4  & 85.9 $\pm$ 3.8 \\
Cable  & 75.2 $\pm$ 5.2&  76.9 $\pm$ 4.1  &  77.6 $\pm$ 3.9 &  75.6 $\pm$ 4.5 &  74.8 $\pm$ 4.1  &  74.7 $\pm$ 3.5 \\
Capsule &   57.9 $\pm$ 7.5 & 60.1 $\pm$ 8.7 &  59.3 $\pm$ 8.4  & 52.1 $\pm$ 6.6  &   58.5 $\pm$ 6.6 & 51.9 $\pm$ 6.6 \\
Carpet  & 64.4 $\pm$ 7.0  & 60.7 $\pm$ 4.1  &  63.9 $\pm$ 6.8  & 52.6 $\pm$ 5.8  &  52.9 $\pm$ 1.8  & 44.9 $\pm$ 1.1 \\
Grid    &  88.1 $\pm$ 7.1 & 83.7 $\pm$ 5.3 &  79.0 $\pm$ 5.9  &  73.9 $\pm$ 7.7 &  65.8 $\pm$ 6.4  &  52.6 $\pm$ 4.5\\
Hazelnut &  80.7 $\pm$ 6.3  & 82.9 $\pm$ 6.7 & 79.3 $\pm$ 11.3  &  80.3 $\pm$ 6.5 &  74.5 $\pm$ 10.6  & 68.5 $\pm$ 11.1 \\
Leather  &  99.2 $\pm$ 0.7 &  99.1 $\pm$ 0.6 & 98.5 $\pm$ 0.5  & 97.7 $\pm$ 1.2 &  95.6 $\pm$ 1.8  & 76.1 $\pm$ 8.5 \\
Metal-nut & 75.3 $\pm$ 7.6  & 75.4 $\pm$ 8.5  &  74.0 $\pm$ 8.4  & 74.9 $\pm$ 7.7 &   75.9 $\pm$ 7.6 &  82.5 $\pm$ 2.6\\
Pill  & 64.8 $\pm$ 6.2  &  65.1 $\pm$ 5.7 &  66.5 $\pm$ 7.0  &  60.5 $\pm$ 7.2 & 56.3 $\pm$ 7.9 & 59.5 $\pm$ 4.5 \\
Screw   &  72.4 $\pm$ 7.7 &  71.7 $\pm$ 9.2  &  75.7 $\pm$ 19.0  &  67.8 $\pm$ 15.5 &  65.9 $\pm$ 15.9 & 41.5 $\pm$ 3.2 \\
Tile   &  83.1 $\pm$ 4.6  & 81.7 $\pm$ 3.9 & 81.4 $\pm$ 6.9  & 78.7 $\pm$ 2.7 & 70.4 $\pm$ 6.4 &  51.7 $\pm$ 4.9\\
Toothbrush  & 61.5 $\pm$ 6.0  & 63.2 $\pm$ 3.6  &  69.5 $\pm$ 7.7  & 59.6 $\pm$ 3.3 & 60.7 $\pm$ 3.9 & 64.1 $\pm$ 4.9 \\
Transistor  & 74.9 $\pm$ 2.0 & 74.8 $\pm$ 3.7 &  79.2 $\pm$ 4.7 &  74.7 $\pm$ 5.6 & 80.3 $\pm$ 5.2 & 82.9 $\pm$ 5.3 \\
Wood  &  94.5 $\pm$ 0.6  &  95.0 $\pm$ 1.2  &  95.8 $\pm$ 1.1 &  94.8 $\pm$ 1.8 & 95.3 $\pm$ 1.1 &  95.7 $\pm$ 1.4 \\
Zipper&   85.6 $\pm$ 4.9 &   81.3 $\pm$ 6.6 &  80.4 $\pm$ 5.9 & 77.3 $\pm$ 6.9  &  77.3 $\pm$ 5.8 &  79.6 $\pm$ 4.7\\

 \midrule
 Avg &  78.0 $\pm$ 5.1  &   77.6 $\pm$ 5.0 & 78.0 $\pm$ 6.7 &  74.1 $\pm$ 5.8  & 73.0 $\pm$ 5.8 &  67.5 $\pm$ 4.7 \\
 \bottomrule \\

 \end{tabular}}

 \caption{ Effect of using a different \textbf{percentage of patches} for defect detection in the \textbf{One-Shot}, \textbf{Five-Shot} and \textbf{Ten-Shot} settings, as described in the main text, Sec.~4.3, Fig.~7.}
 \label{tab:defect_detection_ablation_1shot}
 \end{table*}